\documentclass[lettersize,journal]{IEEEtran}
\usepackage{amsmath,amsfonts}
\usepackage{algorithmic}
\usepackage{array}
\usepackage{textcomp}
\usepackage{stfloats}
\usepackage{verbatim}
\usepackage{graphicx}
\usepackage{balance}
\usepackage{amssymb}
\usepackage{epsfig}
\usepackage{multirow}
\usepackage{tabularx}
\usepackage{adjustbox}
\usepackage{siunitx}
\usepackage{makecell}
\usepackage{stmaryrd} 
\usepackage{latexsym} 
\usepackage{url}
\usepackage{amsfonts}
\usepackage{xkeyval}

\usepackage[breaklinks,colorlinks,bookmarks]{hyperref}
\usepackage{booktabs}

\usepackage{subcaption}

\usepackage{xcolor,colortbl}
\definecolor{Gray1}{gray}{0.85}
\definecolor{Gray2}{gray}{0.65}
\definecolor{darkgreen}{RGB}{0,127,0}
\definecolor{darkred}{RGB}{200,0,0}

\usepackage{pifont}
\usepackage[capitalize]{cleveref}
\crefname{section}{Sec.}{Secs.}
\Crefname{section}{Section}{Sections}
\Crefname{table}{Table}{Tables}
\crefname{table}{Tab.}{Tabs.}

\newcommand{\figref}[1]{Fig.~\ref{#1}}

\newcommand{\tabref}[1]{Tab.~\ref{#1}}

\newcommand{\Figref}[1]{Figure~\ref{#1}}

\newcommand{\Tabref}[1]{Table~\ref{#1}}

\newcommand{\ie}{\textit{i.e.}}
\newcommand{\eg}{\textit{e.g.}}
\newcommand{\etal}{\textit{et al.}}

\definecolor{Aquamarine}{rgb}{0.32, 0.70, 0.73}

\usepackage{bbm}

\begin{document}

\title{All-day Depth Completion via Thermal-LiDAR Fusion}
%
\author{Janghyun Kim$^{1}$, Minseong Kweon$^{1}$, Jinsun Park$^{1*}$, Ukcheol Shin$^{2*}$
\\[1.0em]
$^{1}$Pusan National University \quad $^{2}$Carnegie Mellon University
\thanks{Janghyun Kim is with the Department of Information Convergence Engineering (Artificial Intelligence Major), Pusan National University, Busan, Republic of Korea (e-mail: jangjoa41@pusan.ac.kr)}
\thanks{Minseong Kweon is with the Research Institute of Computers, Information and Communication, Pusan National University, Busan, Republic of Korea (e-mail: wou1202@pusan.ac.kr)}
\thanks{Jinsun Park is with the School of Computer Science and Engineering, Pusan National University, Busan, Republic of Korea (e-mail: jspark@pusan.ac.kr).} 
\thanks{Ukcheol Shin is with the Robotics Institute, Carnegie Mellon University, Pittsburgh, Pennsylvania, United States (e-mail: ushin@andrew.cmu.edu).} 
%
\thanks{$*$ Corresponding author}
}

\maketitle

\begin{abstract}
Depth completion, which estimates dense depth from sparse LiDAR and RGB images, has demonstrated outstanding performance in well-lit conditions.
However, due to the limitations of RGB sensors, existing methods often struggle to achieve reliable performance in harsh environments, such as heavy rain and low-light conditions.
Furthermore, we observe that ground truth depth maps often suffer from large missing measurements in adverse weather conditions such as heavy rain, leading to insufficient supervision.
In contrast, thermal cameras are known for providing clear and reliable visibility in such conditions, yet research on thermal-LiDAR depth completion remains underexplored.
Moreover, the characteristics of thermal images, such as blurriness, low contrast, and noise, bring unclear depth boundary problems.
To address these challenges, we first evaluate the feasibility and robustness of thermal-LiDAR depth completion across diverse lighting (\eg, well-lit, low-light), weather (\eg, clear-sky, rainy), and environment (\eg, indoor, outdoor) conditions, by conducting extensive benchmarks on the MS$^2$ and ViViD datasets.
In addition, we propose a framework that utilizes COntrastive learning and Pseudo-Supervision (COPS) to enhance depth boundary clarity and improve completion accuracy by leveraging a depth foundation model in two key ways.
First, COPS enforces a depth-aware contrastive loss between different depth points by mining positive and negative samples using a monocular depth foundation model to sharpen depth boundaries.
Second, it mitigates the issue of incomplete supervision from ground truth depth maps by leveraging foundation model predictions as dense depth priors.
We also provide in-depth analyses of the key challenges in thermal-LiDAR depth completion to aid in understanding the task and encourage future research.
\end{abstract}


\section{Introduction}
\label{sec:intro}

Depth completion is a critical task in various real-world applications, including autonomous driving~\cite{bai2020depthnet}, robotics~\cite{popovic2021volumetric}, and augmented reality~\cite{syed2022in}. 
Numerous algorithms~\cite{liu2017learning, park2020non, hu2021penet, lin2022dynamic, wang2023lrru} have been developed to estimate dense depth maps by fusing information from RGB and LiDAR sensors. These networks are typically trained and evaluated on well-established datasets such as KITTI Depth Completion (KITTI DC)~\cite{kitti}, DDAD~\cite{guizilini20203d}, and NYUv2~\cite{NYU}, which focus on RGB-based depth completion.
However, networks trained on these datasets often struggle to generalize to all-day scenarios, particularly under low-light conditions or adverse weather.

\begin{figure}[t]
    \centering
    \begin{subfigure}[t]{1.0\linewidth}
        \centering
        \includegraphics[width=0.96\linewidth]{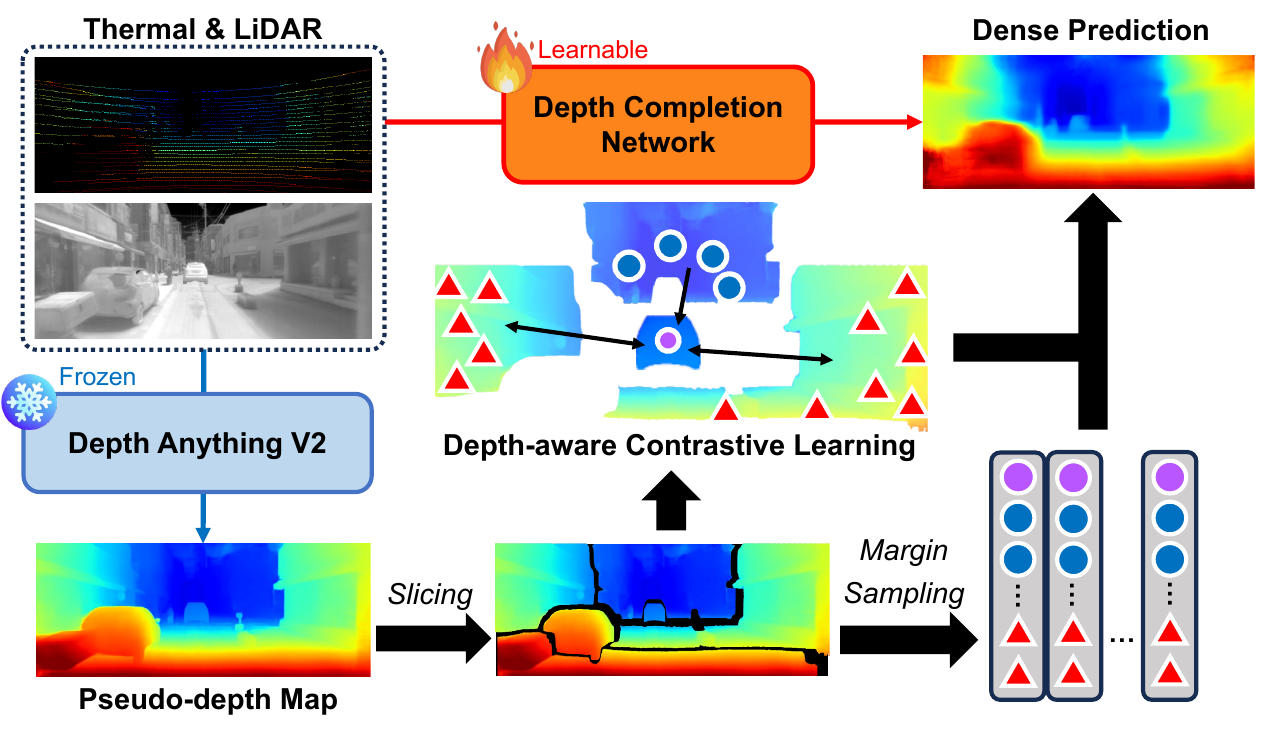} \\ \vspace{-0.01in}
        \caption{\footnotesize{Overview of the proposed depth-aware contrastive learning approach.}}
        \label{fig:teaser_contrastive_learning}
    \end{subfigure}
    \par\smallskip
    \begin{subfigure}[t]{1.0\linewidth}
        \centering
        \renewcommand{\arraystretch}{0.3}
        \begin{tabular}{c@{\hskip 0.0015\linewidth}c@{\hskip 0.0015\linewidth}c}
            \includegraphics[width=0.31\linewidth]{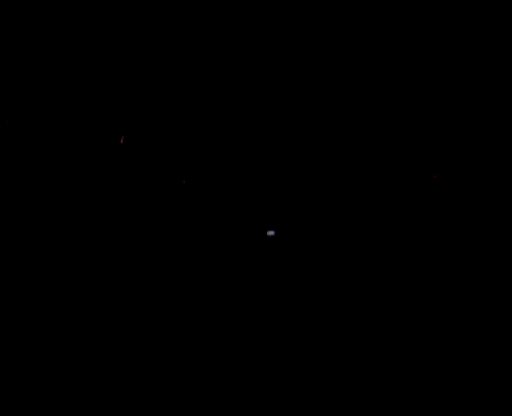} &
            \includegraphics[width=0.31\linewidth]{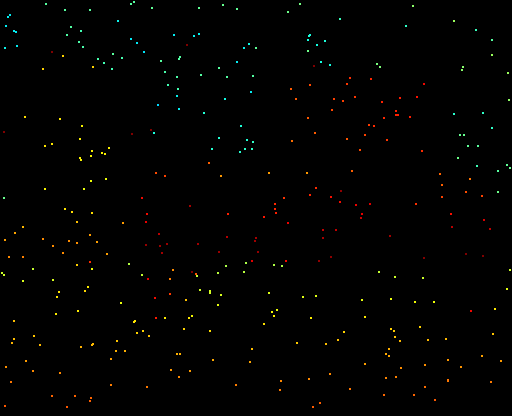} &
            \includegraphics[width=0.31\linewidth]{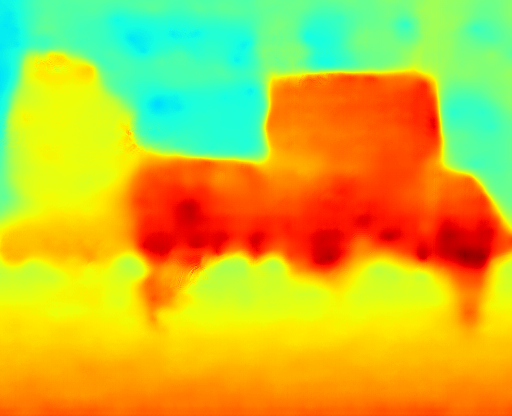} \\
            \footnotesize{RGB Image} & \footnotesize{Sparse Depth} & \footnotesize{LRRU~\cite{wang2023lrru}} \\
            \includegraphics[width=0.31\linewidth]{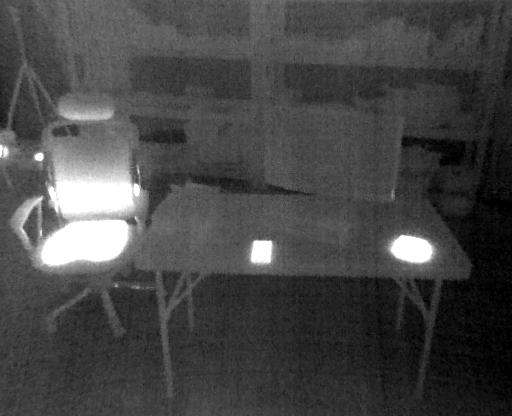} &
            \includegraphics[width=0.31\linewidth]{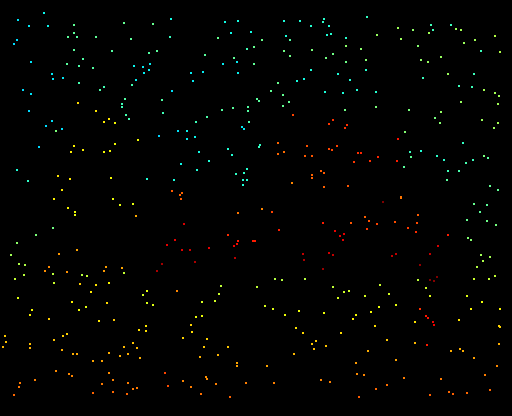} &
            \includegraphics[width=0.31\linewidth]{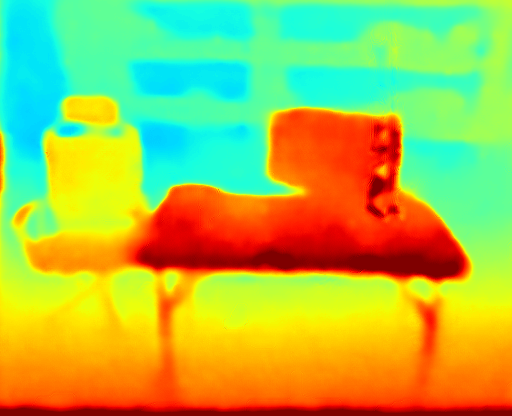} \\
            \footnotesize{Thermal Image} & \footnotesize{Sparse Depth} & \footnotesize{Ours}
        \end{tabular}
        \caption{\footnotesize{Depth completion results using RGB and thermal images.}}
        \label{fig:teaser_pred_comparison}
    \end{subfigure}
    \caption{\textbf{Overview of the proposed method and depth completion result comparison between RGB and thermal modalities.} 
    The proposed contrastive learning method (a) aims to mitigate blurry depth boundaries and insufficient supervision issues caused by thermal images and adverse weather. The qualitative results (b) highlight the significant advantages of thermal-LiDAR fusion in low-light conditions.
    }
    \label{fig:teaser_compare}
\end{figure}

Depth completion in these challenging conditions presents distinct difficulties compared to daytime scenarios, including increased LiDAR sparsity, degraded RGB image quality, and significant variations in lighting and surface reflectivity.
Therefore, depth completion models in such conditions require alternative sensors for robustness and reliability.
As shown in \cref{fig:teaser_pred_comparison}, RGB cameras rely on ambient lighting and perform poorly in low-visibility environments.
In contrast, thermal cameras capture infrared radiation emitted by objects, allowing them to consistently perceive scene structures regardless of external lighting.
As a result, thermal-based depth completion achieves reliable performance in challenging environments, effectively preserving structural details.

Although the robustness of thermal imaging has been widely utilized in various fields, such as depth estimation~\cite{shin2023deep, park2022adaptive, kim2024exploiting}, segmentation~\cite{sun2020fuseseg, kim2021ms}, and object detection~\cite{munir2021sstn,lee2024crossformer}, its application to depth completion remains underexplored.
Therefore, we first benchmark existing RGB-based depth completion methods on thermal images using the Multi-Spectral Stereo ($\text{MS}^2$)~\cite{shin2023deep} and ViViD~\cite{lee2022vivid++} datasets, providing a comprehensive evaluation and analysis of existing algorithms in various harsh environments.
The $\text{MS}^2$ dataset provides RGB, thermal, and LiDAR data from outdoor environments with diverse conditions such as day, night, and rain. 
In contrast, the ViViD dataset offers RGB, thermal, and sparse depth data from indoor scenes, encompassing both well-lit and low-light conditions.

\begin{figure}[t]
    \centering
    \renewcommand{\arraystretch}{0.2}
    \begin{tabular}{c@{\hskip 0.003\linewidth}c}
        \includegraphics[width=0.47\linewidth]{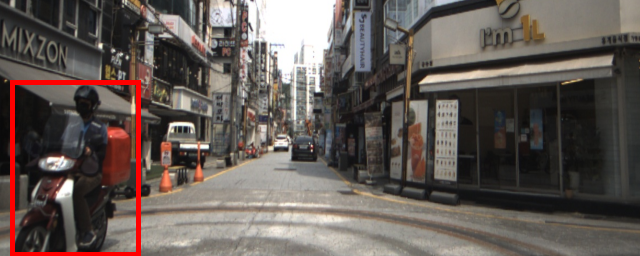}  & 
        \includegraphics[width=0.47\linewidth]{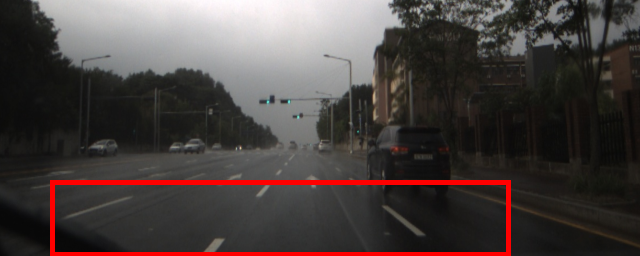} \\
        \includegraphics[width=0.47\linewidth]{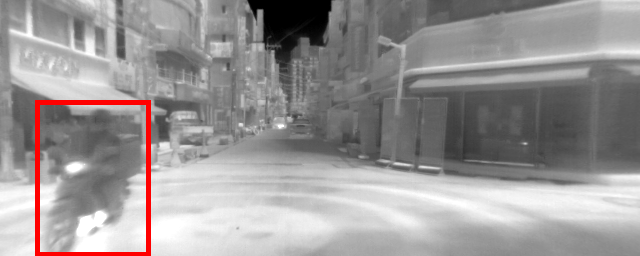}  & 
        \includegraphics[width=0.47\linewidth]{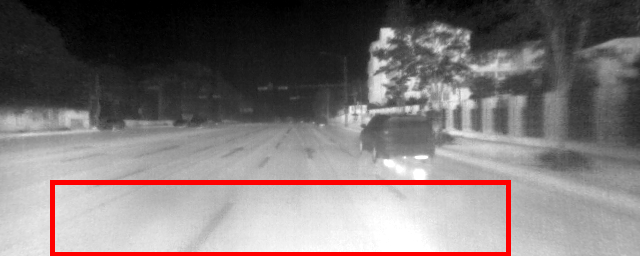} \\
        \includegraphics[width=0.47\linewidth]{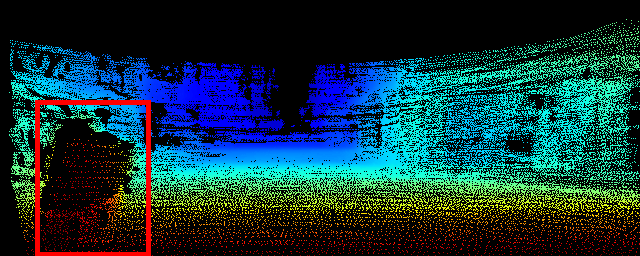}  & 
        \includegraphics[width=0.47\linewidth]{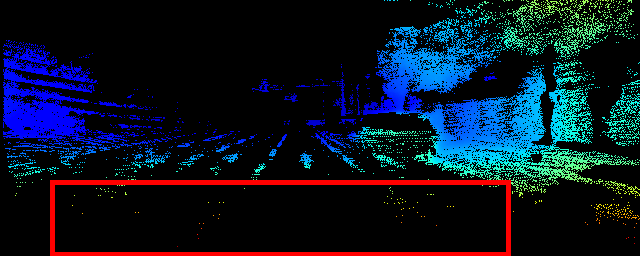} \\
        \footnotesize{$\text{MS}^2$-Day} & \footnotesize{$\text{MS}^2$-Rain}
    \end{tabular}
    \caption{\textbf{Missing LiDAR measurements and blurry thermal image problems in the $\text{MS}^2$ dataset~\cite{shin2023deep}.}}
    \label{fig:empty_lidar}
\end{figure}

After that, to address the challenges of thermal depth completion in harsh environments, we introduce a novel framework that utilizes COntrastive learning and Pseudo-Supervision (COPS) from a depth foundation model.
Specifically, COPS tackles two critical issues in thermal-LiDAR depth completion, as illustrated in \cref{fig:empty_lidar}: (i) unclear depth boundaries in thermal images and (ii) missing ground truth (GT) regions resulting from inherent limitations of LiDAR sensors in adverse conditions like rain or low-light environments.
The first key component of COPS is a depth-aware contrastive learning approach, as shown in \figref{fig:teaser_contrastive_learning}.
%
It aims to distill the sharp depth boundaries produced by foundation models into the target depth completion models by enforcing contrastive loss between different depth points.
The second component is pseudo-supervision, where depth maps estimated from the foundation model serve as pseudo-classes to compensate for missing GT regions.
As a result, COPS not only sharpens depth boundaries but also mitigates the challenges of sparse and incomplete GT regions, significantly improving depth completion performance in real-world environments.
Our contributions can be summarized as follows:
\begin{itemize}
\item 
We evaluate representative depth completion algorithms across RGB and thermal modalities to establish standardized benchmark results on the MS$^2$ and ViViD dataset, which covers diverse real-world scenarios (\eg, low-light and rainy conditions).
\item We propose a novel COntrastive and Pseudo-Supervised learning (COPS) framework that integrates a depth-aware contrastive learning with pseudo-classes to address blurry boundaries and missing GT region issues in thermal-LiDAR depth completion.
\item 
We provide in-depth analyses of the challenges inherent in thermal-LiDAR depth completion and potential future research topics in this field.
\end{itemize}

\section{Related Works}
\label{sec:related works}

\subsection{Depth Completion}
Image-guided depth completion methods~\cite{qiu2019deeplidar, zhao2021adaptive, liu2021fcfr, yan2022rignet, yan2023desnet, zhou2023bev, zhao2023deep, kim2024adnet}  utilize RGB images to generate dense depth maps by refining sparse depth inputs with rich contextual cues from RGB images, such as texture, color, and object boundaries.
A Spatial Propagation Network (SPN) is a representative image-guided completion method that iteratively refines depth maps by leveraging spatial relationships between image pixels.
Various SPN-based methods~\cite{liu2017learning, cheng2018depth, park2020non, lin2022dynamic, wang2023lrru} have been developed to improve depth propagation techniques across the spatial domain.
SPN~\cite{liu2017learning} initially introduced the concept of propagation-based depth completion, which was later improved by CSPN~\cite{cheng2018depth} through recursive depth prediction using fixed affinity values. 
NLSPN~\cite{park2020non} advanced this approach by employing deformable convolution, enabling the network to capture long-range dependencies with relevant affinities. 
DySPN~\cite{lin2022dynamic} utilizes dynamic affinity values during propagation, resulting in more precise depth estimation.

CompletionFormer~\cite{youmin2023completionformer} and PENet~\cite{hu2021penet} adopted these SPN techniques in their final stages to refine the depth map.
For other image-guided methods without SPN, S2D~\cite{ma2018sparse} employs an early-fusion approach to integrate depth and RGB information, while GuideNet~\cite{tang2020learning} enhances feature fusion by leveraging a guided convolutional module across multiple stages, effectively combining image and depth features for improved depth prediction.

Although these works have demonstrated promising results in RGB domain datasets, such as the KITTI Depth Completion dataset~\cite{kitti}, their performance often degrades under challenging conditions, such as nighttime or rainy weather. 
In these adverse scenarios, increased LiDAR sparsity and degraded RGB image quality lead to unreliable and inaccurate prediction results.
To address this limitation, we explore thermal-LiDAR fusion for depth completion task, leveraging the lighting-invariant and environment-robust properties of thermal cameras for robust perception in adverse conditions.

\subsection{Depth Foundation Model}
Earlier monocular depth estimation networks~\cite{fu2018deep,bhat2021adabins} have predominantly concentrated on in-domain metric depth estimation, where both training and test images belong to the same domain. However, the practicality of these methods is often constrained in real-world scenarios, leading to increasing interest in zero-shot relative monocular depth estimation (\ie, depth foundation model). Some methods~\cite{fu2024geowizard, ke2024repurposing} tackle this problem by refining model architectures, such as employing Stable Diffusion~\cite{rombach2022high} as a depth denoising mechanism.

Other approaches adopt a data-centric approach, leveraging large-scale datasets. For instance, MiDaS~\cite{ranftl2020towards} collects 2M labeled images and employs scale-invariant loss to improve generalization. Given the difficulties in scaling labeled datasets, Depth Anything~\cite{yang2024depth_v1} instead utilizes 62M unlabeled images to enhance model robustness. 
Depth Anything V2~\cite{yang2024depth} replaces all labeled real images with synthetic images and scales up the capacity of the teacher model, leading to outstanding depth accuracy and generalization performances.
Here, to leverage its highest performances, we adopt Depth Anything V2 as a foundation depth model for our COPS framework.

\subsection{Contrastive Learning}

Recently, contrastive learning has demonstrated significant improvements across various vision tasks~\cite{he2020momentum, chen2020improved} by learning discriminative feature representations. Unlike conventional supervision, contrastive learning methods enforce similar samples to be closer in the embedding space while pushing dissimilar ones apart.
These approaches have shown promise in tasks involving class-specific relationships, such as classification~\cite{caron2020unsupervised, chen2020simple}, object detection~\cite{xie2021detco}, and semantic segmentation~\cite{khosla2020supervised, zhao2021contrastive, wang2021exploring}. 
In image classification, SimCLR~\cite{chen2020simple} aims to learn general representations through data augmentation, focusing on capturing a broad feature space. SwAV~\cite{caron2020unsupervised} refines class distribution boundaries by combining enhanced augmentations with online clustering to improve class separation.
For object detection, DetCo~\cite{xie2021detco} improves both image-level and instance-level feature learning by incorporating multi-scale features and combining global and local contrastive learning.

In semantic segmentation, contrastive learning has been employed to enhance feature discrimination and effectively delineate class boundaries.
Zhao~\etal~\cite{zhao2021contrastive} demonstrated that contrastive learning helps enforce intra-class compactness and inter-class separability, leading to better segmentation performance.
Wang~\etal~\cite{wang2021exploring} further introduced a pixel-wise contrastive framework that refines segmentation boundaries by leveraging a hard sampling strategy based on semantic class labels. Specifically, they employed hardest sampling for challenging positives and negatives, semi-hard sampling for balanced training, and segmentation-aware hard anchor sampling to refine misclassified boundaries.
This pixel-level contrastive learning and sample mining strategies have demonstrated its potential for dense prediction tasks.
In these works, sample mining methods are typically guided by class information, using semantic labels or image-level class distinctions.

However, the lack of explicit labels in the depth completion task makes it difficult to establish clear criteria for distinguishing positive and negative pairs.
Therefore, we propose a depth-aware contrastive learning approach that selects the pairs based on pseudo-depth values at the pixel level.
By leveraging pseudo-depth maps from a frozen monocular depth foundation model, our method enhances feature discrimination, distinguishing depth boundaries and improving overall prediction quality. 
\section{Method} 
\label{sec:method}
\begin{figure*}[t]
    \centering
\includegraphics[width=1.00\linewidth]{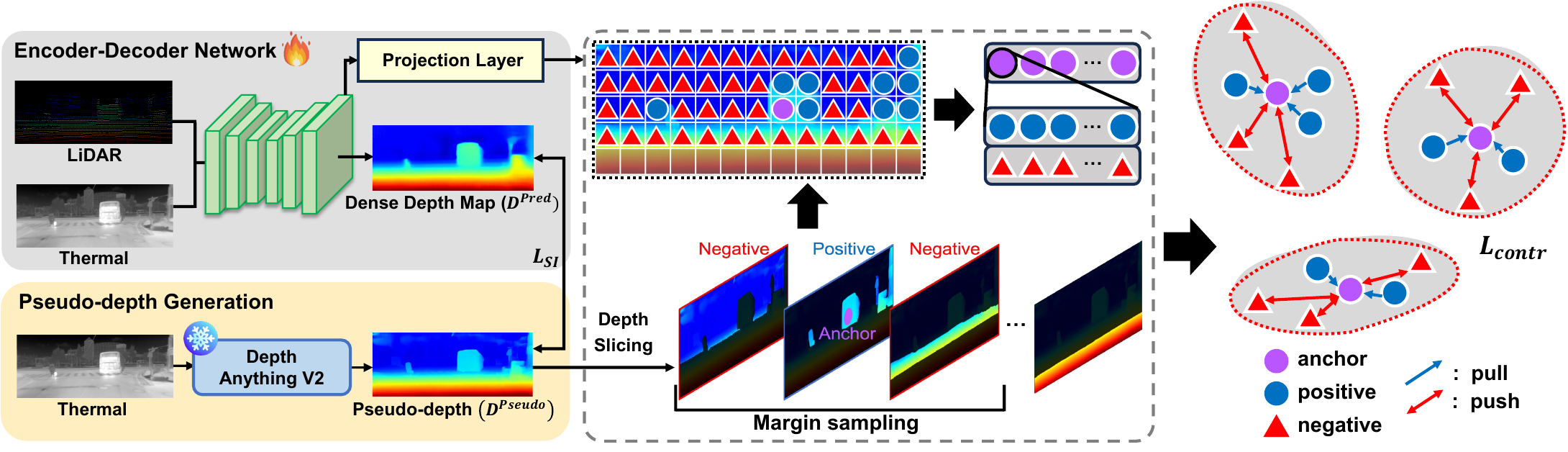}
    \center
    \caption{\textbf{Overall framework of our depth completion.} Our encoder-decoder network takes thermal image and LiDAR points as input, while pseudo-depth generation module only utilizes thermal image. The network is directly supervised using the pseudo-depth map and further incorporates it as a contrastive learning criterion through depth slicing.}
    \label{fig:Overall_Architecture}
\end{figure*}

In this section, we propose a novel and first framework designed for the thermal-LiDAR depth completion task. 
The proposed framework utilizes contrastive learning and pseudo-supervision to address missing GT problems in adverse weather conditions and blurry boundary problems in thermal images, respectively.

\subsection{Overall Framework}
\label{sec:overview}
\Cref{fig:Overall_Architecture} illustrates the overall framework of our thermal depth completion algorithm.
Our framework is designed to be able to seamlessly integrate various existing depth completion networks (\eg, NLSPN~\cite{park2020non}, GuideNet~\cite{tang2020learning}, and LRRU~\cite{wang2023lrru}).
In detail, our framework consists of the following three modules.

\subsubsection{Depth Completion Network}
\label{subsec:Depth Completion Network}
We utilize an encoder-decoder network to generate a dense depth map from a thermal image and a sparse depth map, with supervision from ground truth data, following the standard depth completion paradigm.
Our approach is designed to preserve the original architecture of existing depth completion networks while improving the overall depth map quality by leveraging a pseudo-depth map.

\subsubsection{Pseudo-depth Generation}
\label{subsec:Pseudo-depth Generation}
The pseudo-depth map is derived from a frozen monocular depth foundation model and utilized it into  COntrastive and Pseudo-Supervised
learning (COPS) strategy to enhance robustness during the training process. We selected Depth Anything V2~\cite{yang2024depth} as the depth foundation model due to its demonstrated high accuracy across various datasets and its ability to provide reliable absolute depth scale estimates.

\subsubsection{COPS}
\label{subsec:COPS}
We introduce a depth-aware contrastive learning strategy to enhance feature representation, leading to a sharp depth boundary. 
Specifically, we slice a pseudo-depth map to assign indices (\ie, classes) to each depth range.
After that, we sample positive and negative samples by considering minimum and maximum margins.
For a given anchor point, pixels within a minimum margin are assigned as positive samples.
Conversely, pixels falling between the minimum and maximum margins, often considered as confusing depth values, are treated as negative samples.
We refine the representation space through our contrastive learning to have a clear distribution boundary for each depth range.
Furthermore, our framework is directly supervised using scale-invariant loss with a pseudo-depth map to effectively compensate for the lack of dense ground truth depth.
Notably, our framework requires no additional computation during inference compared to a standard encoder-decoder network.

\subsection{Depth Completion Network}
\label{sec:Depth_completion_network}
Depth completion algorithms typically follow an encoder-decoder architecture, where a sparse depth map and an image are used as inputs to generate a dense depth prediction.
In our approach, both RGB and thermal depth completion use an image and sparse depth as inputs to estimate a dense depth map. Since thermal images are inherently single-channel, we adapt the original encoder-decoder structure by replicating the thermal channel to match the standard three-channel input format.
We employ each method as originally designed, whether it includes Spatial Propagation Network (SPN)~\cite{cheng2018depth,park2020non} or not. The SPN module refines the final depth output by predicting affinity and offset using guidance features, while other approaches estimate the depth map directly from the final decoder features, following their respective architectures.

\subsection{Depth-aware Contrastive Learning}
\label{sec:Depth-aware Contrastive Learning}

\subsubsection{Pseudo-class Generation}
\label{subsec:Pseudo-class Generation}
We determine positive and negative sample pairs based on pseudo-depth values, leveraging the reliability of pseudo-depth in capturing fine-grained structures and dense depth information.
Unlike ground truth depth, which is often sparse and may lack detailed edge information, pseudo-depth from a foundation model offers a dense and consistent representation, making it well-suited for defining meaningful pseudo-classes in contrastive learning.
For this purpose, we discretize the depth range to assign pseudo-classes for each pixel.
Assume that we discretize a depth range $\left[d_{min}, d_{max}\right]$ into $M$ intervals.
Then, the discretized depth range is given by $\left\{d_{min}, d_{min}+\Delta d, \cdots, d_{min}+i\Delta d, \cdots, d_{max} \right\}$ where $\Delta d = \frac{d_{max}-d_{min}}{M}$.
The pseudo-class $y_{j}$ of the pixel $j$ is defined as follows:
\begin{equation}
    y_{j} = i \qquad \mathrm{if} ~~~ d_{i} \leq D^{Pseudo}_{j} < d_{i+1},
    \label{eq:pseudo-class}
\end{equation}
where $D^{Pseudo}_{j}$ is the pseudo-depth value at pixel $j$ and $d_{i} = d_{min} + i \Delta d$.
Note that we set $d_{min} = 0$, $d_{max}$ to the maximum pseudo-depth value across all pixels, and $M=2d_{max}$.
This configuration yields an interval width of $\Delta d = 0.5$.

\subsubsection{Margin Sampling}
\label{subsec:Margin_Sampling}
Based on these pseudo-classes, we further introduce a margin sampling method to focus on near-negative samples within a specific range of depth differences.
For a query pixel $q$ with pseudo-class $y_q$, the positive and negative sample sets are defined as:
\begin{gather}
    \mathcal{P}(q) = \{ k \mid |y_q - y_k| < \psi_{\text{min}} \}, \label{eq:margin_sampling_positive}
    \\
    \mathcal{N}(q) = \{ k \mid \psi_{\text{min}} \leq |y_q - y_k| \leq \psi_{\text{max}} \}. \label{eq:margin_sampling_negative}
\end{gather}
Here, each element $k$ denotes the index of a reference pixel. The set of $\mathcal{P}(q)$ contains indices of positive samples, and $\mathcal{N}(q)$ contains indices of negative samples.
The term $|y_q - y_k|$ represents the absolute difference between the pseudo-classes of the query pixel and a sample pixel.
The parameters $\psi_\text{min}$ and $\psi_\text{max}$ specify the minimum and maximum margins, respectively, with $\psi_\text{min}=1$ in our method.
This negative sample mining (\ie, $\mathcal{N}(q)$) encourages the model to focus on challenging cases where depth variations are subtle but crucial for accurate depth estimation.
Furthermore, we limit the maximum number of samples per pseudo-class $i$ to $n$ to maintain class balance and reduce computational complexity when constructing positive and negative sample sets. This ensures that the sampling process remains efficient while preserving a balanced representation of positive and negative samples across pseudo-classes.

\subsubsection{Contrastive Learning Loss}
\label{subsec:Contrastive Learning Loss}
To align feature representations with the pseudo-depth map, we extract a feature $f$ from the last decoder using a 1$\times$1 convolutional projection layer.
Since both the guidance feature for the SPN module and the final decoder feature used for depth regression without SPN play a crucial role in preserving depth boundary sharpness and prediction accuracy, we refine these features by aligning the representation space through contrastive learning.
We compute a self-similarity matrix $s(q, k) = {f_{q}} \cdot {f_{k}}^{T}$, following the previous approach~\cite{wang2021exploring}, where each element represents the similarity relationship between the pixels $q$ and $k$.

Our proposed contrastive loss function with margin sampling is given by:
\begin{gather}
    \mathcal{L}_{contr} = \sum_{q \in Q} \Bigl( \frac{1}{|\mathcal{P}(q)|} \sum_{k^{+} \in \mathcal{P}(q)} 
    - \log \frac{\exp \left( \frac{s(q, k^{+})}{\tau} \right)}{Z_q} \Bigr),
    \\
    Z_q = \exp \left( \frac{s(q, k^{+})}{\tau}  \right) + \sum_{k^{-} \in \mathcal{N}(q)} \exp \left( \frac{s(q, k^{-})}{\tau}  \right),
\label{eq:contr_loss}
\end{gather}
where $Q$ is the set of query pixels, $k^{+}$ and $k^{-}$ denote positive and negative samples, $s(q, k^{+})$ and $s(q, k^{-})$ represent the positive and negative similarity scores, and $\tau$ is the temperature parameter controlling the sharpness of the softmax function.
The $\mathcal{N}(q)$ is selected based on a margin-based sampling strategy in \cref{eq:margin_sampling_negative}.
By excluding distant negatives that contribute less to the learning process, the margin sampling method allows for more effective contrastive learning, leading to finer feature discrimination and depth prediction.

To further refine the contrastive learning process, we exclude pixels that have corresponding points in the ground truth data. This ensures the model focuses on meaningful feature discrimination derived from pseudo-depth predictions, rather than relying on fully supervised data. By avoiding ground truth regions, we mitigate bias and enable the model to better capture nuanced relationships among unsupervised pixels.

\subsection{Pseudo-depth Supervision}
\label{sec:Pseudo-depth Supervision}
We adopt a straightforward self-supervision approach by employing the pseudo-depth map with a scale-invariant loss~\cite{eigen2014depth}.
The scale-invariant loss was originally introduced for monocular depth estimation tasks~\cite{bhat2021adabins, bhat2023zoedepth, yang2024depth} and has demonstrated its effectiveness in addressing varying depth scales.
Although Depth Anything V2 provides metric depth (\ie, absolute depth), the distribution of depth values in the image can exhibit slight discrepancies.

Therefore, we employ the scale-invariant loss to provide supervision that is independent of scale, eliminating the need for fine-tuning the depth foundation model.
We also introduce a random sampling strategy to handle depth variations without being overly biased toward specific regions.
This method effectively fills the empty regions in sparse LiDAR points, while ensuring a more balanced and robust depth prediction across the entire image.
Our pseudo-depth supervision loss is defined as follows:
\begin{equation}
    L_{SI} = \frac{1}{|\mathcal{R}_{\text{sample}}|} \left( \sum_{j \in \mathcal{R}_{\text{sample}}} d_{j}^{2} - \lambda \cdot \left( \sum_{j \in \mathcal{R}_{\text{sample}}} d_{j} \right)^{2} \right), 
    \label{eq:scale-invariant}
\end{equation}
where $d_{j} = \log D^{Pseudo}_{j} - \log D^{Pred}_{j}$. Here, $D^{Pseudo}_{j}$ and $D^{Pred}_{j}$ represent the pseudo-depth value and the predicted depth value at pixel $j$, respectively.
The set $\mathcal{R}_{\text{sample}}$ comprises a randomly selected subset of pixels based on the sampling ratio $\alpha \in (0,1]$.
We set $\lambda = 0.5$, following another mono depth estimation models~\cite{yang2024depth}. This value effectively balances the scale-invariant term and the mean bias term, ensuring that both local depth variations and global consistency are accurately captured.

\subsection{Loss Functions}
\label{sec:Loss Functions}
\subsubsection{Depth Completion Loss}  
\label{subsec:Depth_Completion_Loss}
To ensure a fair comparison of all the existing algorithms for benchmarking, we utilize the same loss function as follows:
\begin{equation}
    \mathcal{L}_{base} =  w_{e}\mathcal{L}_e +w_{GT}\mathcal{L}_{GT},
\label{eq:base_loss}
\end{equation}
where $\mathcal{L}_e$ and $\mathcal{L}_{GT}$ denote smooth edge loss and smooth L1 loss, respectively.
The smooth edge loss \( \mathcal{L}_e \) is formulated as:
\begin{equation}
    \mathcal{L}_e = \left| \partial_x D^{Pred} \right| e^{-\left| \partial_x I \right|} + 
    \left| \partial_y D^{Pred} \right| e^{-\left| \partial_y I \right|},
\label{eq:smooth_loss}
\end{equation}
where $D^{Pred}$ represents the predicted depth map, and \( I \) denotes the corresponding intensity image. This loss encourages spatial smoothness in depth predictions while preserving edge details guided by the intensity image.
The smooth L1 loss \( \mathcal{L}_{GT} \) is defined as follows:
\begin{gather}
    \mathcal{L}_{GT} = \frac{1}{N} \sum_{j=1}^{N} l_j,
\\
    l_j = 
    \begin{cases} 
        \frac{1}{2} \left( D^{GT}_{j} - D^{Pred}_{j} \right)^2, & \text{if } \left| D^{GT}_{j} - D^{Pred}_{j} \right| < 1, \\
        \left| D^{GT}_{j} - D^{Pred}_{j} \right| - \frac{1}{2}, & \text{otherwise.}
    \end{cases}
\label{eq:l1_loss}
\end{gather}
Here, $D^{GT}_{j}$ and $D^{Pred}_{j}$ represent the ground truth depth and predicted depth map at pixel \( j \), respectively, and \( N \) is the total number of pixels.

\subsubsection{Contrastive and Pseudo-supervised Loss}
\label{subsec:Contrastive_Loss}
In addition to the baseline loss, we adopt our proposed supervision approaches to several networks. To address potential conflicts between the two supervision signals, we further introduce a stage-learning strategy defined as follows:
\begin{equation}
L_{pseudo} = \beta L_{SI} + (1 - \beta) L_{contr},
\label{eq:pseudo_loss}
\end{equation}
where $\beta = \mathbbm{1}_{\{t \leq T/2\}}$. Here, $\mathbbm{1}$ represents the indicator function, which returns 1 if the condition $t \leq T/2$ holds and 0 otherwise. $T$ represents the total number of epochs and $t$ is the current epoch.
We first apply supervision using the scale-invariant loss $L_{SI}$ during the first half of the training process to focus on global consistency. This ensures that the network learns a robust understanding of the overall depth structure and relationships within the scene.
During the latter half of the training, we switch to using the contrastive loss ($L_{contr}$) to refine local consistency. This supervision emphasizes sorting relative depth relationships, helping the model capture fine-grained depth details.
By decoupling two supervisions, the model achieves improved depth-aware representations, balancing both large-scale scene understanding and detailed local features.
Finally, the overall loss function of our framework is defined as follows:
\begin{equation}
    \mathcal{L}_{total} = \mathcal{L}_{base} + w_{pseudo}\mathcal{L}_{pseudo}.
\label{eq:total_loss}
\end{equation}
\section{Experimental Results}
\label{sec:experiments}

\begin{table*}[t!]
\caption{Depth completion results on the evaluation set of $\text{MS}^2$ dataset~\cite{shin2023deep}.}
\centering
\footnotesize
\resizebox{0.92\linewidth}{!}{
\begin{tabular}{c|c|c|c|c|c|c|c|c|c}
\Xhline{3\arrayrulewidth}
\multirow{3}{*}{Methods}  & \multirow{3}{*}{TestSet} & \multicolumn{4}{c|}{\textbf{Thermal + LiDAR}} & \multicolumn{4}{c}{\textbf{RGB + LiDAR}}
\\ \cline{3-10}
& & RMSE  & MAE  & iRMSE  & iMAE & RMSE  & MAE  & iRMSE  & iMAE \\
& & (m)  & (m)  & (1/mm)  & (1/mm) & (m)  & (m)  & (1/mm)  & (1/mm) \\ \hline\hline
\multirow{4}{*}{CSPN~\cite{cheng2020cspn}} & Day  & 2.456  & 1.278  & 7.566  & 4.229  & 2.848  & 1.569 & 8.236 & 4.572 \\
& Night  & 2.558  & 1.488  & 7.631  & 5.081  & 2.882 & 1.745 & 7.674 & 5.011 \\
& Rain  & 2.956  & 1.730  & 9.445  & 5.661  & 4.532 & 2.763 & 11.330 & 6.548 \\
& \cellcolor{Gray1}Average  & \cellcolor{Gray1}2.664  & \cellcolor{Gray1}1.504  & \cellcolor{Gray1}8.225  & \cellcolor{Gray1}4.997  & \cellcolor{Gray1}3.450  & \cellcolor{Gray1}2.045 & \cellcolor{Gray1}9.142 & \cellcolor{Gray1}5.407 \\ \hline
\multirow{4}{*}{S2D~\cite{ma2018sparse}} & Day  & 2.418  & 1.232  & 6.655  & 3.942  & 2.805  & 1.492  & 81.684  & 4.591 \\
& Night  & 2.480  & 1.440  & 7.177  & 4.885  & 2.858  & 1.670  & 55.648  & 4.668 \\
& Rain  & 2.868  & 1.640  & 8.772  & 5.167  & 4.428  & 2.671  & 87.356  & 6.926 \\
& \cellcolor{Gray1}Average  & \cellcolor{Gray1}2.596  & \cellcolor{Gray1}1.442  & \cellcolor{Gray1}7.531  & \cellcolor{Gray1}4.674  & \cellcolor{Gray1}3.392  & \cellcolor{Gray1}1.963 & \cellcolor{Gray1}75.303 & \cellcolor{Gray1}5.436 \\ \hline
\multirow{4}{*}{NLSPN~\cite{park2020non}} & Day  & 2.133  & 1.091  & 5.981  & 3.540  & 2.871  & 1.532  & 7.118  & 4.012 \\
& Night  & 2.366  & 1.389  & 6.847  & 4.702  & 2.857  & 1.691  & 6.913  & 4.528 \\
& Rain  & 2.569  & 1.507  & 7.546  & 4.677  & 4.596  & 2.762  & 10.420  & 6.401 \\
& \cellcolor{Gray1}Average  & \cellcolor{Gray1}2.361  & \cellcolor{Gray1}1.333  & \cellcolor{Gray1}6.797  & \cellcolor{Gray1}4.320  & \cellcolor{Gray1}3.472  & \cellcolor{Gray1}2.015  & \cellcolor{Gray1}8.212  & \cellcolor{Gray1}5.017 \\ \hline
\multirow{4}{*}{GuideNet~\cite{tang2020learning}} & Day  & 2.092  & 1.077  & 28.960  & 3.895  & 2.853  & 1.535  & 17.937  & 4.427 \\
& Night  & 2.276  & 1.330  & 17.532  & 4.782  & 2.774  & 1.660  & 11.815  & 4.676 \\
& Rain  & 2.566  & 1.467  & 14.241  & 4.758  & 4.593  & 2.761  & 25.751  & 6.861 \\
& \cellcolor{Gray1}Average  & \cellcolor{Gray1}2.318  & \cellcolor{Gray1}1.295  & \cellcolor{Gray1}20.130  & \cellcolor{Gray1}4.489  & \cellcolor{Gray1}3.439  & \cellcolor{Gray1}2.006  & \cellcolor{Gray1}18.712 & \cellcolor{Gray1}5.362 \\ \hline
\multirow{4}{*}{DySPN~\cite{lin2022dynamic}} & Day  & 2.091  & 1.080  & 5.879  & 3.437  & 2.892  & 1.495  & 6.959  & 3.834 \\
& Night  & 2.295  & 1.351  & 6.696  & 4.068  & 3.071  & 1.744  & 6.962 & 4.477 \\
& Rain  & 2.511  & 1.441  & 6.928  & 4.238  & 4.456  & 2.632  & 10.258  & 6.298 \\
& \cellcolor{Gray1}Average  & \cellcolor{Gray1}\underline{2.304}  & \cellcolor{Gray1}\underline{1.294}  & \cellcolor{Gray1}\underline{6.510}  & \cellcolor{Gray1}\textbf{3.921}  & \cellcolor{Gray1}3.499  & \cellcolor{Gray1}1.974  & \cellcolor{Gray1}8.119  & \cellcolor{Gray1}4.906 \\ \hline
\multirow{4}{*}{CompletionFormer~\cite{youmin2023completionformer}} & Day  & 2.113  & 1.087  & 6.081  & 3.586  & 2.899 & 1.505  & 9.946  & 3.903 \\
& Night  & 2.315  & 1.353  & 6.752  & 4.644  & 2.24  & 1.658  & 6.850  & 4.296 \\
& Rain  & 2.497  & 1.448  & 7.496  & 4.559  & 4.351  & 2.615  & 17.401  & 6.716 \\
& \cellcolor{Gray1}Average  & \cellcolor{Gray1}2.313  & \cellcolor{Gray1}1.299  & \cellcolor{Gray1}6.780  & \cellcolor{Gray1}4.276  & \cellcolor{Gray1}3.417  & \cellcolor{Gray1}1.944  & \cellcolor{Gray1}11.569  & \cellcolor{Gray1}5.017 \\ \hline
\multirow{4}{*}{LRRU~\cite{wang2023lrru}} & Day  & 2.087  & 1.080  & 5.570  & 3.326  & 2.758  & 1.441  & 6.889  & 3.857 \\
& Night  & 2.270  & 1.342  & 6.627  & 4.584  & 2.677  & 1.558  & 6.545  & 4.264 \\
& Rain  & 2.485  & 1.439  & 6.842  & 4.282  & 4.376  & 2.617  & 9.861  & 5.995 \\
& \cellcolor{Gray1}Average  & \cellcolor{Gray1}\textbf{2.286} & \cellcolor{Gray1}\textbf{1.290}  & \cellcolor{Gray1}\textbf{6.357}  & \cellcolor{Gray1}\underline{4.066}  & \cellcolor{Gray1}\underline{3.300}  & \cellcolor{Gray1}\underline{1.892}  & \cellcolor{Gray1}\textbf{7.822}  & \cellcolor{Gray1}\underline{4.739} \\ \hline
\multirow{4}{*}{BP-Net~\cite{tang2024bilateral}} & Day  & 2.207  & 1.120  & 5.775  & 3.442  & 2.639  & 1.350  & 6.623  & 3.650 \\
& Night  & 2.350  & 1.367  & 6.696  & 4.601  & 2.840  & 1.662  & 6.578  & 4.300 \\
& Rain  & 2.671  & 1.531  & 7.239  & 4.438  & 4.165  & 2.504  & 10.147  & 5.958 \\
& \cellcolor{Gray1}Average  & \cellcolor{Gray1}2.416  & \cellcolor{Gray1}1.344  & \cellcolor{Gray1}6.585  & \cellcolor{Gray1}4.165  & \cellcolor{Gray1}\textbf{3.240}  & \cellcolor{Gray1}\textbf{1.843}  & \cellcolor{Gray1}\underline{7.846}  & \cellcolor{Gray1}\textbf{4.670} \\
\Xhline{3\arrayrulewidth}
\multicolumn{10}{r}{\textbf{Bold}: The best, \underline{Underline}: The second-best} \\
\end{tabular}}
\label{tab:MS2_result}
\caption{Depth completion results on the evaluation set of ViViD dataset~\cite{lee2022vivid++}.}
\footnotesize
\centering
\resizebox{0.92\linewidth}{!}{
\begin{tabular}{c|c|c|c|c|c|c|c|c|c}
\Xhline{3\arrayrulewidth}
\multirow{3}{*}{Methods}  & \multirow{3}{*}{TestSet} & \multicolumn{4}{c|}{\textbf{Thermal + LiDAR}} & \multicolumn{4}{c}{\textbf{RGB + LiDAR}} \\ \cline{3-10}
& & RMSE  & MAE  & iRMSE  & iMAE & RMSE  & MAE  & iRMSE  & iMAE \\
& & (m)  & (m)  & (1/m)  & (1/m) & (m)  & (m)  & (1/m)  & (1/m) \\ \hline\hline
\multirow{3}{*}{CSPN~\cite{cheng2020cspn}} 
& Indoor-bright  & 0.201  & 0.099  & 0.070 & 0.015  & 0.289  & 0.143 & 2.119 & 0.071 \\ 
& Indoor-dark  & 0.194  & 0.096  & 0.066 & 0.015  & 0.374 & 0.224 & 2.142 & 0.084 \\ 
& \cellcolor{Gray1}Average  & \cellcolor{Gray1}0.195  & \cellcolor{Gray1}0.097 & \cellcolor{Gray1}0.067 & \cellcolor{Gray1}0.015 & \cellcolor{Gray1}0.345 & \cellcolor{Gray1}0.201 & \cellcolor{Gray1}\textbf{2.136} & \cellcolor{Gray1}0.080 \\ \hline
\multirow{3}{*}{S2D~\cite{ma2018sparse}} 
& Indoor-bright  & 0.204  & 0.108  & 0.071  & 0.016  & 0.303  & 0.157  & 2.120  & 0.075 \\ 
& Indoor-dark  & 0.195  & 0.104  & 0.066  & 0.016  & 0.357  & 0.225  & 2.142  & 0.082 \\ 
& \cellcolor{Gray1}Average  & \cellcolor{Gray1}0.198  & \cellcolor{Gray1}0.105  & \cellcolor{Gray1}0.067  & \cellcolor{Gray1}0.016  & \cellcolor{Gray1}0.344  & \cellcolor{Gray1}0.208  & \cellcolor{Gray1}\underline{2.137}  & \cellcolor{Gray1}0.081 \\ \hline
\multirow{3}{*}{NLSPN~\cite{park2020non}} 
& Indoor-bright  & 0.194  & 0.091  & 0.070  & 0.013  & 0.270  & 0.121  & 2.119  & 0.069 \\ 
& Indoor-dark  & 0.185  & 0.086  & 0.066  & 0.014  & 0.327  & 0.180  & 2.142  & 0.073 \\ 
& \cellcolor{Gray1}Average  & \cellcolor{Gray1}0.188  & \cellcolor{Gray1}0.087  & \cellcolor{Gray1}0.067  & \cellcolor{Gray1}0.014  & \cellcolor{Gray1}0.312  & \cellcolor{Gray1}0.165  & \cellcolor{Gray1}\textbf{2.136}  & \cellcolor{Gray1}0.072 \\ \hline
\multirow{3}{*}{GuideNet~\cite{tang2020learning}} 
& Indoor-bright  & 0.191  & 0.089  & 0.069  & 0.013  & 0.281  & 0.120  & 2.119  & 0.070 \\ 
& Indoor-dark  & 0.185  & 0.085  & 0.064  & 0.013  & 0.339  & 0.188  & 2.142  & 0.075 \\ 
& \cellcolor{Gray1}Average  & \cellcolor{Gray1}\underline{0.186}  & \cellcolor{Gray1}0.086  & \cellcolor{Gray1}\underline{0.065}  & \cellcolor{Gray1}\underline{0.013}  & \cellcolor{Gray1}0.324  & \cellcolor{Gray1}0.171  & \cellcolor{Gray1}\textbf{2.136}  & \cellcolor{Gray1}0.074 \\ \hline
\multirow{3}{*}{DySPN~\cite{lin2022dynamic}} 
& Indoor-bright  & 0.202  & 0.088  & 0.070  & 0.012  & 0.276  & 0.112  & 2.119  & 0.067 \\ 
& Indoor-dark  & 0.194  & 0.081  & 0.065  & 0.012  & 0.308  & 0.141  & 2.141  & 0.069 \\ 
& \cellcolor{Gray1}Average  & \cellcolor{Gray1}0.196  & \cellcolor{Gray1}\textbf{0.083}  & \cellcolor{Gray1}0.066  & \cellcolor{Gray1} \textbf{0.012}  & \cellcolor{Gray1}0.300  & \cellcolor{Gray1}\underline{0.134}  & \cellcolor{Gray1}\underline{2.137}  & \cellcolor{Gray1}\textbf{0.068} \\ \hline
\multirow{3}{*}{CompletionFormer~\cite{youmin2023completionformer}} 
& Indoor-bright  & 0.189  & 0.093  & 0.068  & 0.013  & 0.278  & 0.109  & 2.119  & 0.068 \\ 
& Indoor-dark  & 0.180  & 0.088  & 0.063  & 0.014  & 0.302  & 0.140  & 2.142  & 0.069 \\ 
& \cellcolor{Gray1}Average  & \cellcolor{Gray1}\textbf{0.182}  & \cellcolor{Gray1}0.089  & \cellcolor{Gray1}\textbf{0.064}  & \cellcolor{Gray1}0.014  & \cellcolor{Gray1}\textbf{0.297}  & \cellcolor{Gray1}\textbf{0.131}  & \cellcolor{Gray1}\underline{2.137} & \cellcolor{Gray1}\underline{0.069} \\ \hline
\multirow{3}{*}{LRRU~\cite{wang2023lrru}} 
& Indoor-bright  & 0.194  & 0.088  & 0.070  & 0.013  & 0.282  & 0.125  & 2.119  & 0.070 \\ 
& Indoor-dark  & 0.184  & 0.083  & 0.065  & 0.013  & 0.303  & 0.155  & 2.142  & 0.072 \\ 
& \cellcolor{Gray1}Average  & \cellcolor{Gray1}0.187  & \cellcolor{Gray1}\underline{0.084}  & \cellcolor{Gray1}0.066  & \cellcolor{Gray1}\underline{0.013}  & \cellcolor{Gray1}\underline{0.298}  & \cellcolor{Gray1}0.147  & \cellcolor{Gray1}\textbf{2.136}  & \cellcolor{Gray1}0.071 \\ 
\Xhline{3\arrayrulewidth}
\multicolumn{10}{r}{\textbf{Bold}: The best, \underline{Underline}: The second-best} \\
\end{tabular}
}
\label{tab:vivid_result}
\end{table*}

In this section, we present the benchmarking datasets and provide a detailed explanation of our framework implementation.
In addition, we analyze the performance of existing depth completion methods applied to thermal and RGB cameras in conjunction with LiDAR sensors under diverse conditions. 
Specifically, we conduct comprehensive experiments with established approaches, including SPN-based networks (\ie, CSPN~\cite{cheng2020cspn}, NLSPN~\cite{park2020non}, DySPN~\cite{lin2022dynamic}, CompletionFormer~\cite{youmin2023completionformer}, LRRU~\cite{wang2023lrru}, and BP-Net~\cite{tang2024bilateral}), and other image-guided networks without SPN (\ie,
S2D~\cite{ma2018sparse} and GuideNet~\cite{tang2020learning}).
Furthermore, we perform detailed ablation studies of our proposed COPS and demonstrate its integration into existing networks to enhance robustness in thermal depth completion task.

\begin{figure*}[t]
    \centering
    \renewcommand{\arraystretch}{0.2}
    \begin{tabular}{c@{\hskip 0.002\linewidth}c@{\hskip 0.002\linewidth}c@{\hskip 0.002\linewidth}c@{\hskip 0.002\linewidth}c@{\hskip 0.002\linewidth}c}
        \includegraphics[width=0.162\linewidth]{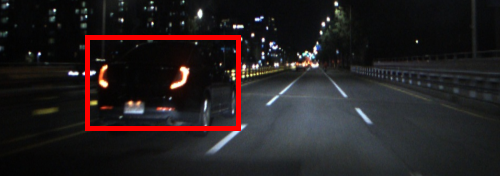} & 
        \includegraphics[width=0.162\linewidth]{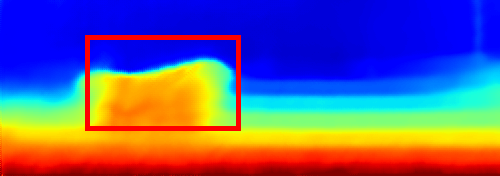} & 
        \includegraphics[width=0.162\linewidth]{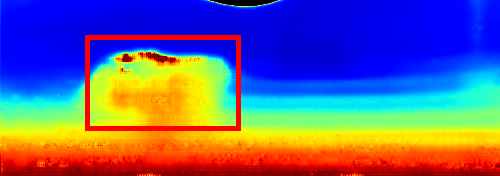} & 
        \includegraphics[width=0.162\linewidth]{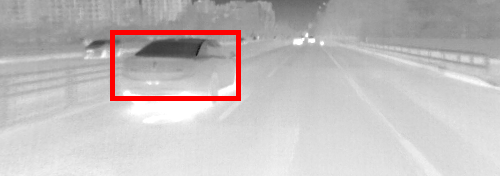} &
        \includegraphics[width=0.162\linewidth]{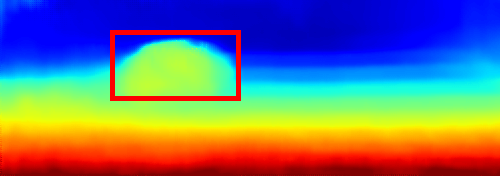} &
        \includegraphics[width=0.162\linewidth]{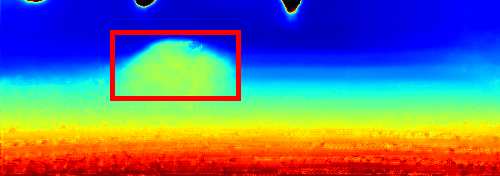} \\
        \includegraphics[width=0.162\linewidth]{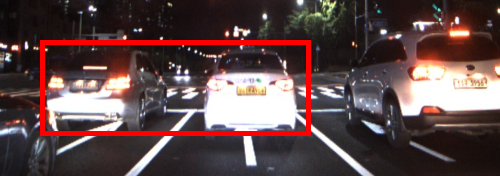} & 
        \includegraphics[width=0.162\linewidth]{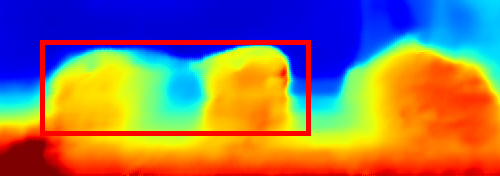} & 
        \includegraphics[width=0.162\linewidth]{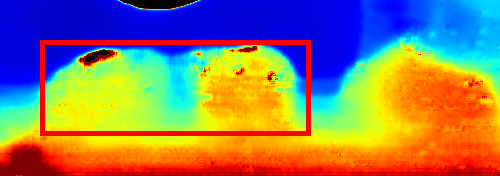} & 
        \includegraphics[width=0.162\linewidth]{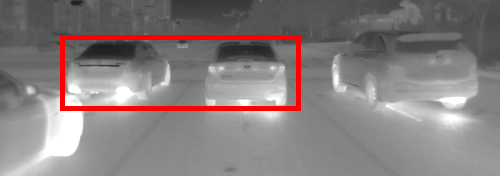} &
        \includegraphics[width=0.162\linewidth]{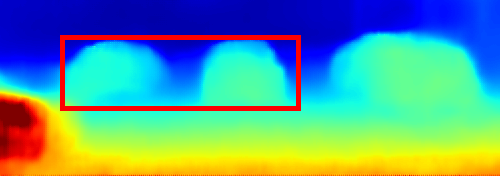} &
        \includegraphics[width=0.162\linewidth]{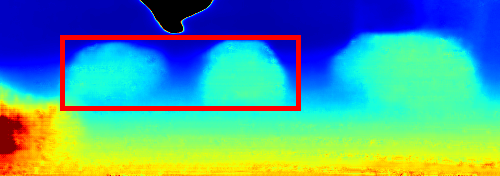} \\
        \includegraphics[width=0.162\linewidth]{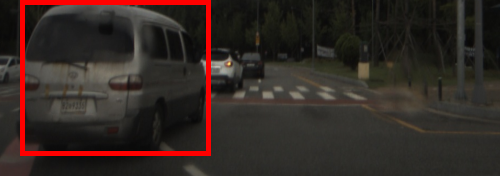} & 
        \includegraphics[width=0.162\linewidth]{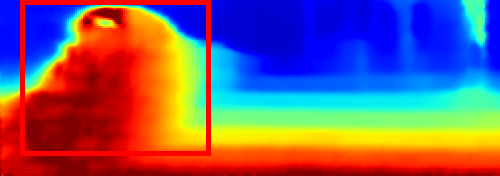} & 
        \includegraphics[width=0.162\linewidth]{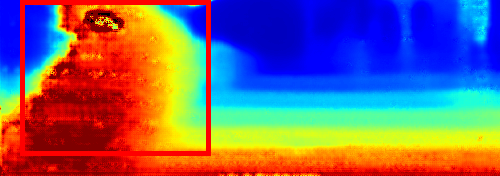} & 
        \includegraphics[width=0.162\linewidth]{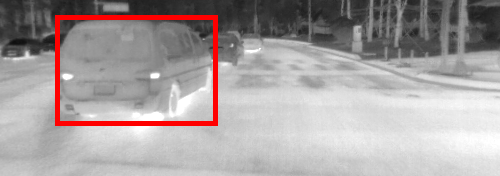} &
        \includegraphics[width=0.162\linewidth]{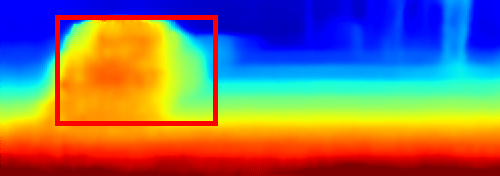} &
        \includegraphics[width=0.162\linewidth]{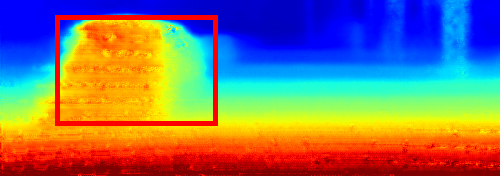} \\
        \includegraphics[width=0.162\linewidth]{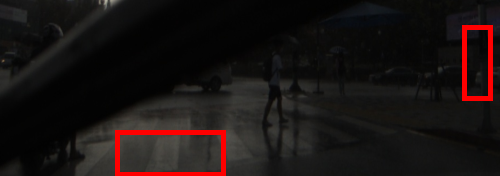} & 
        \includegraphics[width=0.162\linewidth]{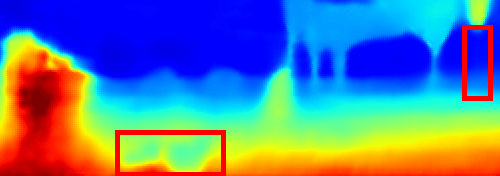} & 
        \includegraphics[width=0.162\linewidth]{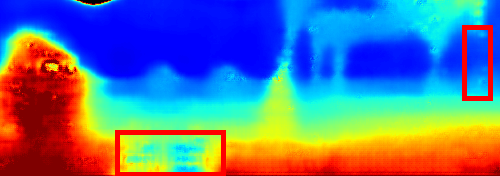} & 
        \includegraphics[width=0.162\linewidth]{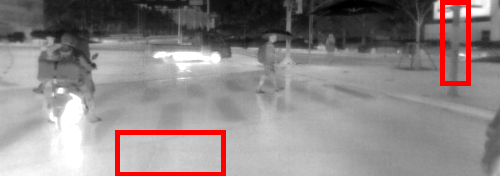} &
        \includegraphics[width=0.162\linewidth]{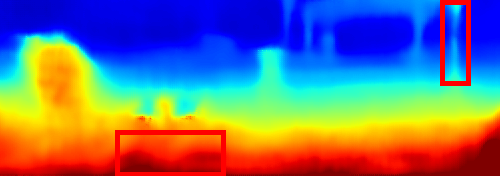} &
        \includegraphics[width=0.162\linewidth]{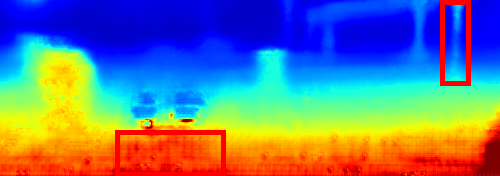} \\
        
        \footnotesize RGB & \footnotesize NLSPN (RGB) & \footnotesize GuideNet (RGB) & \footnotesize Thermal & \footnotesize NLSPN (Thermal) & \footnotesize GuideNet (Thermal)
    \end{tabular}
\caption{
\textbf{Depth map comparisons between two modalities on NLSPN~\cite{park2020non} and GuideNet~\cite{tang2020learning}.}
The first and second rows present the results of nighttime scenarios, while the third and fourth rows correspond to rainy scenarios.}
\label{fig:MS2_modality_compare}
\end{figure*}

\subsection{Implementation Details}
\label{subsec:impl_detail}

\subsubsection{Datasets}
\label{subsubsec:Datasets}
We utilize the Multi-Spectral Stereo ($\text{MS}^2$) dataset~\cite{shin2023deep} as our outdoor benchmark due to its extensive scale and the inclusion of long-wave infrared (\ie, thermal) and RGB cameras.
This dataset provides diverse temporal and weather conditions, including daytime, nighttime, and rainy scenarios, making it well-suited for evaluating thermal depth completion in real-world environments. 
Specifically, the dataset contains over 26K RGB-LiDAR and thermal-LiDAR image pairs for training, 4K for validation, and 2.3K, 2.2K, and 2.5K pairs for evaluation of daytime, nighttime, and rainy conditions, respectively. 
To ensure fair comparisons across modalities, we crop the training images to a resolution of 640$\times$256, while inference is conducted on images at their original resolution.

We select the ViViD dataset~\cite{lee2022vivid++} as our indoor benchmark to evaluate and compare depth completion networks under varying lighting conditions. This dataset includes both indoor-bright and indoor-dark scenarios, allowing for a comprehensive assessment of model performance in low-light environments.
Specifically, the dataset comprises over 2.3K RGB, thermal, and sparse depth data pairs for training, along with 0.2K pairs for validation. For evaluation, we construct a dataset with 0.4K samples under indoor-bright conditions and 1.2K samples under indoor-dark conditions.
Following previous works~\cite{cheng2018depth, park2020non}, we randomly sample 500 depth points per ground truth depth image, as was done on the NYUv2 dataset~\cite{NYU}.
Additionally, we crop the edge boundaries during the inference step, resulting in a shape of 416 × 512 to align with the ground truth of RGB data, while the training data retains its original dimensions.

\subsubsection{Evaluation metric}
\label{subsubsec:Evaluation_metric}
We utilize the following commonly used metrics for depth completion~\cite{cheng2020cspn, park2020non}:
\begin{equation}
\text{Metrics} =
\begin{cases}
\text{RMSE} = \sqrt{\frac{1}{N} \sum\limits_{i=1}^{N} \left( D_i^{\text{GT}} - D_i^{\text{Pred}} \right)^2}, \\
\text{MAE} = \frac{1}{N} \sum\limits_{i=1}^{N} \left| D_i^{\text{GT}} - D_i^{\text{Pred}} \right|, \\
\text{iRMSE} = \sqrt{\frac{1}{N} \sum\limits_{i=1}^{N} \left( \frac{1}{D_i^{\text{GT}}} - \frac{1}{D_i^{\text{Pred}}} \right)^2}, \\
\text{iMAE} = \frac{1}{N} \sum\limits_{i=1}^{N} \left| \frac{1}{D_i^{\text{GT}}} - \frac{1}{D_i^{\text{Pred}}} \right|.
\end{cases}
\end{equation}
We differentiate the metric scale between two datasets in terms of iRMSE and iMAE to highlight the variations in modalities and the adaptability of our method.

\subsubsection{Environments}
\label{subsubsec:Environments}
We conduct all the experiments under consistent conditions. Throughout the experiments, depth completion networks and the proposed methods are trained with a batch size of 12 on the $\text{MS}^2$ dataset for 40 epochs and 50 epochs on the ViViD dataset. The implementation employs the ADAM optimizer and Cosine Annealing Warm Restarts learning rate scheduler using PyTorch Lightning and CUDA 12.1, running on 4 RTX A6000 GPUs.
We set $w_{e} = 0.01$, $w_{GT} = 1.0$, and $w_{pseudo} = 0.2$ for both of datasets.
Moreover, we apply data augmentation techniques using random center crop-and-resize, brightness jitter, horizontal flip, and contrast jitter for all model training.
For thermal image, we incorporate image-wise clipping to adjust temperature values at the image level, group-wise clipping to normalize temperature distributions within grouped regions, and group-wise rearrangement to redistribute temperature values.

\subsection{$\text{MS}^2$ Dataset}
\label{subsec:MS2 Dataset}
\Tabref{tab:MS2_result} provides a detailed comparison of depth completion performance using RGB and thermal modalities on the $\text{MS}^2$ dataset, highlighting distinct trends under challenging conditions such as night and rain. 
While RGB-based depth completion networks often excel in well-lit daytime scenarios, their performance degrades significantly in low-light and adverse weather conditions due to glare, noise, and reduced visibility.
In contrast, thermal-based depth completion consistently outperforms RGB-based methods across various conditions, benefiting from the inherent robustness of thermal imaging against lighting variations and environmental disturbances during training.
For instance, NLSPN~\cite{park2020non} and GuideNet~\cite{tang2020learning} show about 44\% RMSE improvement in rain scenarios and roughly 17\% RMSE improvement in night scenarios when using thermal data compared to RGB.
\Figref{fig:MS2_modality_compare} further highlights the advantages of thermal imaging. In nighttime scenes, depth predictions using RGB cameras are severely impacted by noise caused by movement and lighting variations, whereas thermal cameras produce more stable and consistent depth estimations.
In rainy conditions, RGB-based methods struggle with artifacts introduced by objects such as wipers or water droplets. In contrast, thermal sensors are less affected by surface reflections and provide more accurate depth estimates for both near and far distances.
 
Although BP-Net~\cite{tang2024bilateral} achieves state-of-the-art average performance in RGB-based depth completion, its performance in the thermal domain is hindered by extremely slow convergence. This limitation arises because the pre-processing step of BP-Net are specifically tailored to RGB images.
On the other hand, LRRU~\cite{wang2023lrru} establishes itself as a state-of-the-art network for the thermal modality, achieving competitive performance in the RGB modality with the second-best results in both RMSE and MAE. This success is due to its effective approach to generating the initial depth map~\cite{ku2018defense} and propagating short to long-range distances.
Furthermore, the average performance ranking of thermal imaging networks closely aligns with their ranking on the KITTI Depth Completion dataset~\cite{kitti} except for BP-Net, further demonstrating the stability of thermal depth completion across diverse conditions. 
This consistency makes thermal depth completion a more suitable choice for outdoor applications, effectively overcoming RGB limitations.

\begin{figure*}[t!]
    \centering
    \renewcommand{\arraystretch}{0.2}
    \begin{tabular}{c@{\hskip 0.002\linewidth}c@{\hskip 0.002\linewidth}c@{\hskip 0.002\linewidth}c@{\hskip 0.002\linewidth}c@{\hskip 0.002\linewidth}}
        \includegraphics[width=0.194\linewidth]{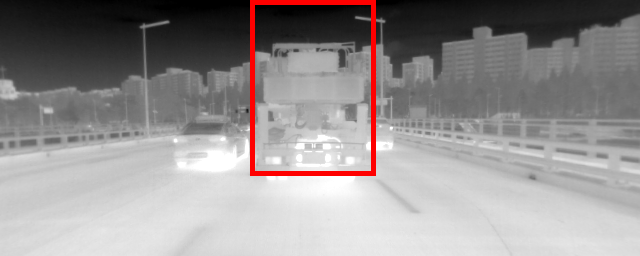} & 
        \includegraphics[width=0.194\linewidth]{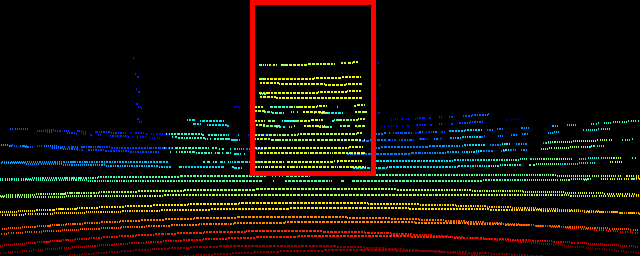} & 
        \includegraphics[width=0.194\linewidth]{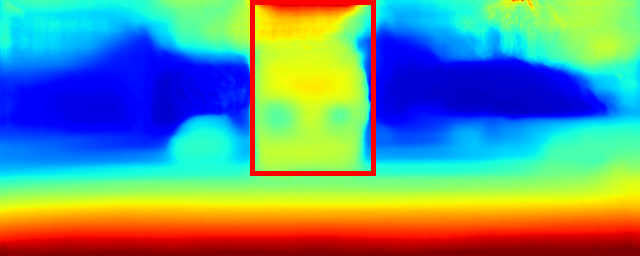} & 
        \includegraphics[width=0.194\linewidth]{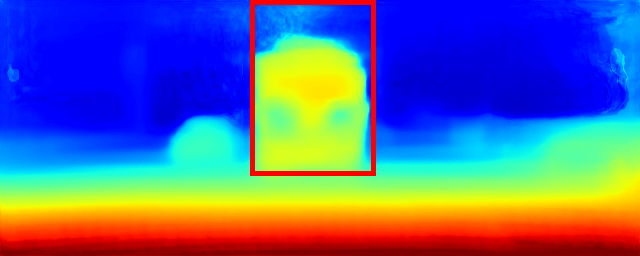} & 
        \includegraphics[width=0.194\linewidth]{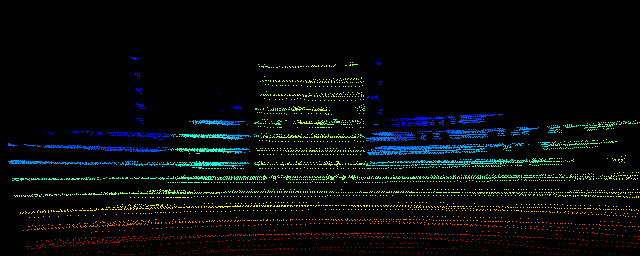} \\
        \includegraphics[width=0.194\linewidth]{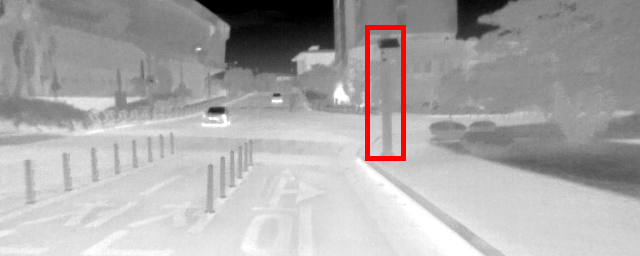} & 
        \includegraphics[width=0.194\linewidth]{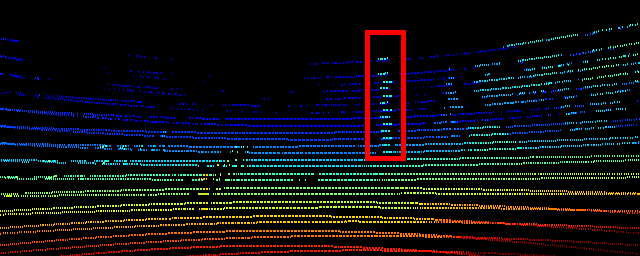} & 
        \includegraphics[width=0.194\linewidth]{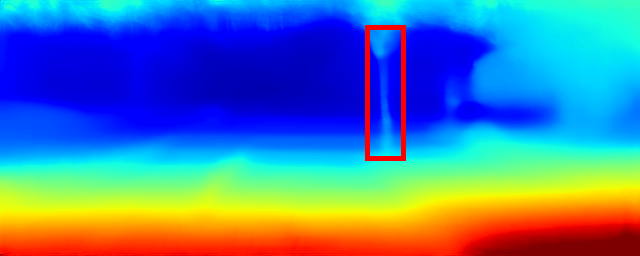} & 
        \includegraphics[width=0.194\linewidth]{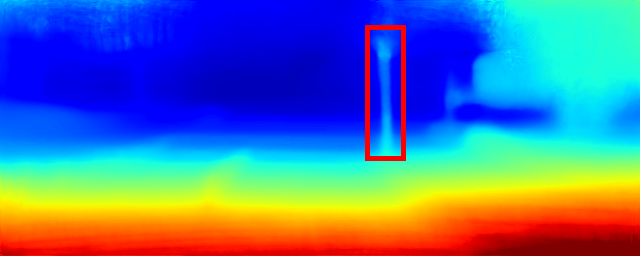} & 
        \includegraphics[width=0.194\linewidth]{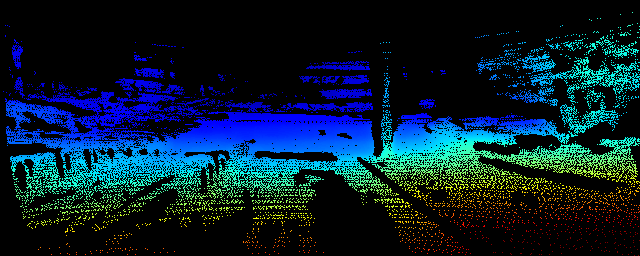} \\
        \includegraphics[width=0.194\linewidth]{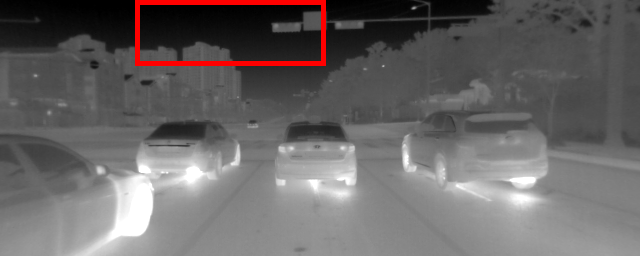} & 
        \includegraphics[width=0.194\linewidth]{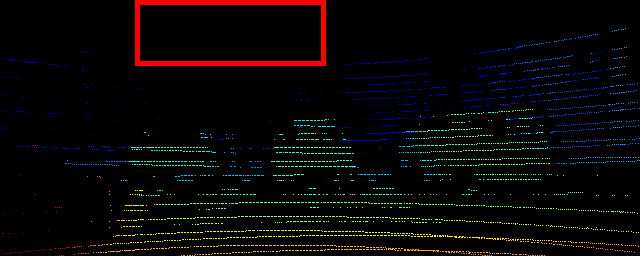} & 
        \includegraphics[width=0.194\linewidth]{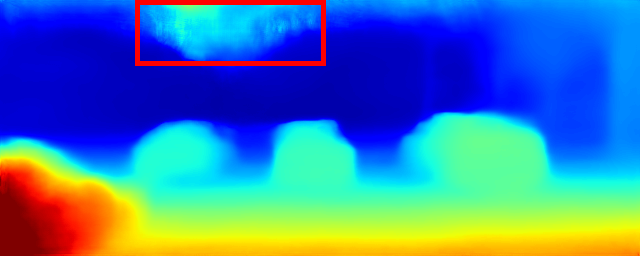} & 
        \includegraphics[width=0.194\linewidth]{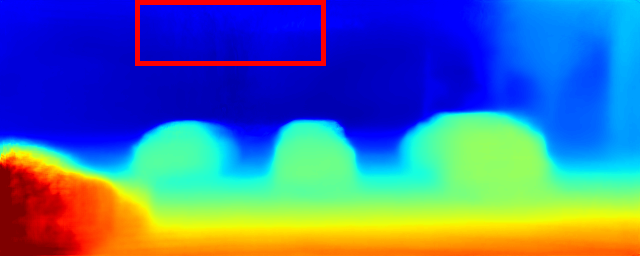} & 
        \includegraphics[width=0.194\linewidth]{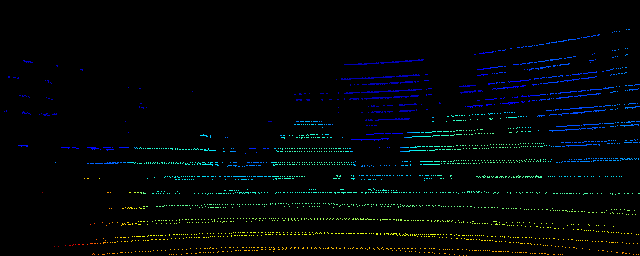} \\
        \includegraphics[width=0.194\linewidth]{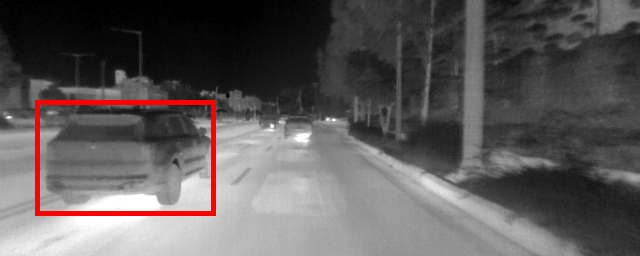} & 
        \includegraphics[width=0.194\linewidth]{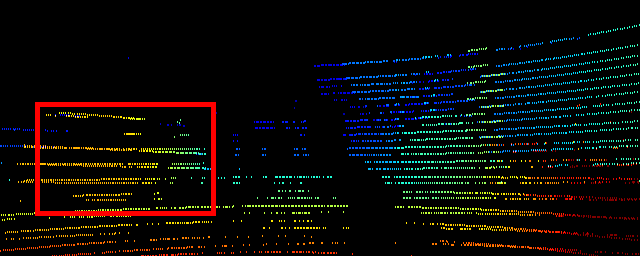} & 
        \includegraphics[width=0.194\linewidth]{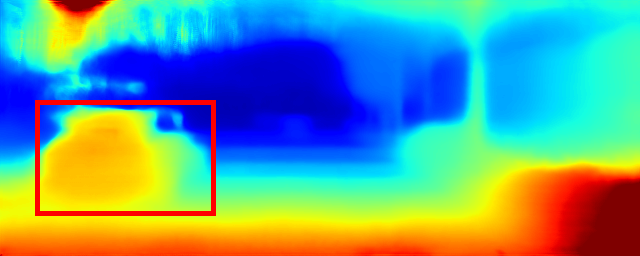} & 
        \includegraphics[width=0.194\linewidth]{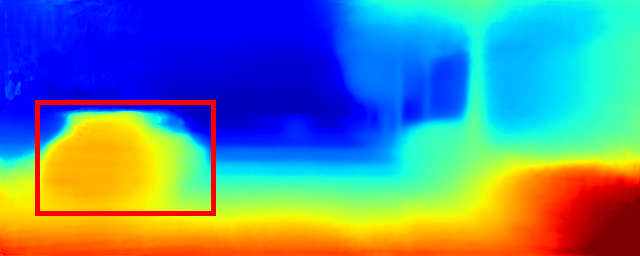} & 
        \includegraphics[width=0.194\linewidth]{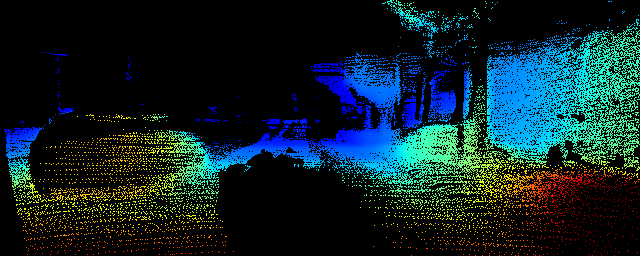} \\
        
        \footnotesize Thermal & \footnotesize Sparse Depth & \footnotesize LRRU~\cite{wang2023lrru} & \footnotesize LRRU+ours & \footnotesize GT
    \end{tabular}
\caption{
\textbf{Depth map comparisons on the $\text{MS}^2$ dataset~\cite{shin2023deep}.} Note that we dilated the sparse depth map.
}
\label{fig:MS2_visual_compare}
\end{figure*}

\begin{table*}[t!]
\caption{Performance comparison of our proposed approach integrated with NLSPN~\cite{park2020non}, GuideNet~\cite{tang2020learning}, and LRRU~\cite{wang2023lrru} on the $\text{MS}^2$~\cite{shin2023deep} and ViViD~\cite{lee2022vivid++} datasets.}
\footnotesize
\centering
\resizebox{0.99\linewidth}{!}{
\begin{tabular}{c|c|cccc|cccc}
\Xhline{3\arrayrulewidth}
\multirow{6}{*}{$\text{MS}^2$} & \multirow{3}{*}{Methods} & \multicolumn{4}{c|}{\textbf{Baseline}} & \multicolumn{4}{c}{\textbf{Ours}} \\ \cline{3-10}
& & RMSE & MAE & iRMSE & iMAE & RMSE & MAE & iRMSE & iMAE \\
& & (m)  & (m)  & (1/mm) & (1/mm) & (m)  & (m)  & (1/mm) & (1/mm) \\ \cline{2-10}
& NLSPN   & 2.361 & 1.333 & 6.797 & 4.320 & 2.347\,(\textcolor{red}{-0.014}) & 1.328\,(\textcolor{red}{-0.005}) & 6.757\,(\textcolor{red}{-0.040}) & 4.285\,(\textcolor{red}{-0.035}) \\
& GuideNet& 2.318 & 1.295 & 20.130 & 4.489 & 2.295\,(\textcolor{red}{-0.023}) & 1.297\,(\textcolor{blue}{+0.002}) & 6.917\,(\textcolor{red}{-13.213}) & 4.398\,(\textcolor{red}{-0.091}) \\
& LRRU    & 2.286 & 1.290 & 6.357  & 4.066 & 2.236\,(\textcolor{red}{-0.050}) & 1.250\,(\textcolor{red}{-0.040}) & 6.201\,(\textcolor{red}{-0.156}) & 3.983\,(\textcolor{red}{-0.083}) \\ \Xhline{2\arrayrulewidth}
\multirow{6}{*}{ViViD} & \multirow{3}{*}{Methods} & \multicolumn{4}{c|}{\textbf{Baseline}} & \multicolumn{4}{c}{\textbf{Ours}} \\ \cline{3-10}
& & RMSE & MAE & iRMSE & iMAE & RMSE & MAE & iRMSE & iMAE \\
& & (m)  & (m)  & (1/m) & (1/m) & (m)  & (m)  & (1/m) & (1/m) \\ \cline{2-10}
& NLSPN   & 0.188 & 0.087 & 0.067 & 0.014 & 0.186\,(\textcolor{red}{-0.002}) & 0.084\,(\textcolor{red}{-0.003}) & 0.066\,(\textcolor{red}{-0.001}) & 0.014\,(\textcolor{black}{0.000}) \\
& GuideNet& 0.186 & 0.086 & 0.065 & 0.013 & 0.182\,(\textcolor{red}{-0.004}) & 0.084\,(\textcolor{red}{-0.002}) & 0.072\,(\textcolor{blue}{+0.007}) & 0.013\,(\textcolor{black}{0.000}) \\
& LRRU    & 0.187 & 0.084 & 0.066 & 0.013 & 0.178\,(\textcolor{red}{-0.009}) & 0.073\,(\textcolor{red}{-0.011}) & 0.064\,(\textcolor{red}{-0.002}) & 0.011\,(\textcolor{red}{-0.002}) \\ \Xhline{3\arrayrulewidth}
\multicolumn{10}{r}{\textcolor{red}{-: Performance improvements}, \textcolor{blue}{+: Performance degradation}} \\
\end{tabular}}
\label{tab:ours}
\end{table*}

\Tabref{tab:ours} demonstrates our supervision approach with depth foundation model achieves superior performance across various networks on outdoor dataset, without introducing additional computation during inference.
Specifically, integrating our method consistently achieves superior performance in terms of RMSE, iRMSE, and iMAE across the NLSPN, GuideNet, and LRRU networks.
Although GuideNet experiences a slight performance degradation in the MAE, the iRMSE stabilizes from 20.130 to 6.917, making it comparable to other depth completion networks.
Furthermore, applying our supervision approach to the LRRU network results in a significant performance improvement, reducing RMSE from 2.286 to 2.236, with a 3.2\% enhancement in MAE.

Our method based on LRRU network effectively address challenges in completing aerial regions and capturing fine-detailed objects in outdoor scenarios, as illustrated in \figref{fig:MS2_visual_compare}.
Despite the advantages of thermal imaging, the absence of LiDAR measurements in nighttime and rainy scenes results in blurred depth predictions for thin structures such as poles and vehicles in the LRRU network.
However, by leveraging a depth foundation model extracted from thermal images and using it as a contrastive learning prior for feature discrimination, our method effectively differentiate depth variations and preserve fine details, particularly near object boundaries where depth discontinuities are critical.
Furthermore, the incorrect predictions in aerial regions are significantly mitigated through direct supervision from the pseudo-depth map.
Since the depth foundation model generates a dense pseudo-depth map independent of LiDAR data, it serves as a reliable reference, allowing the network to learn meaningful depth representations even in upper regions where conventional depth completion methods often struggle.

\begin{figure*}[t]
    \centering
    \renewcommand{\arraystretch}{0.2}
    \begin{tabular}{@{}c@{\hskip 0.003\linewidth}c@{\hskip 0.003\linewidth}c@{\hskip 0.003\linewidth}c@{\hskip 0.003\linewidth}c@{\hskip 0.003\linewidth}c@{\hskip 0.003\linewidth}c}
        \includegraphics[width=0.14\linewidth]{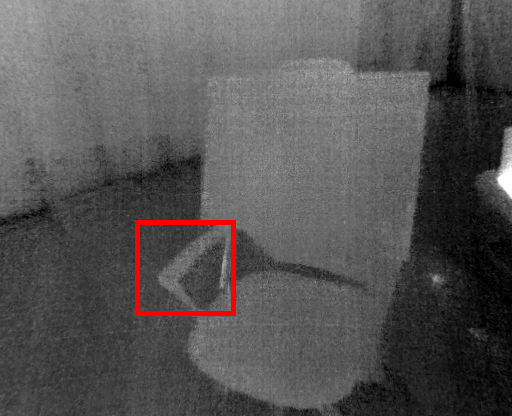} & 
        \includegraphics[width=0.14\linewidth]{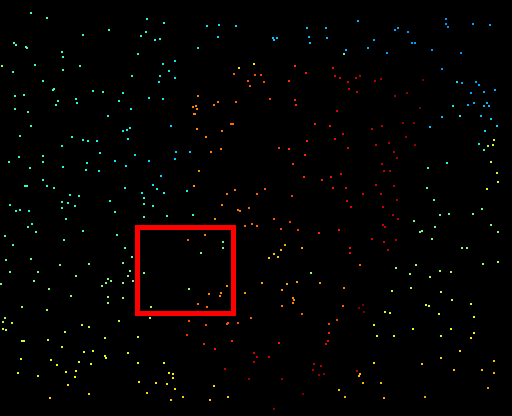} & 
        \includegraphics[width=0.14\linewidth]{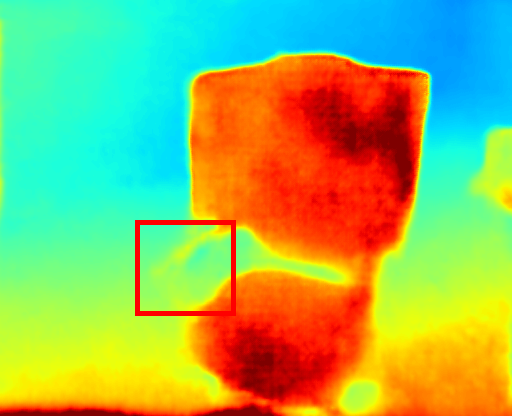} & 
        \includegraphics[width=0.14\linewidth]{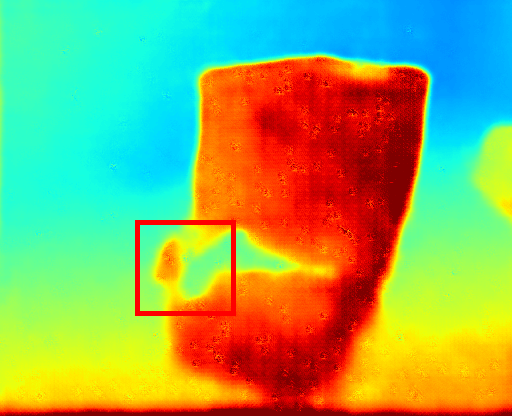} &
        \includegraphics[width=0.14\linewidth]{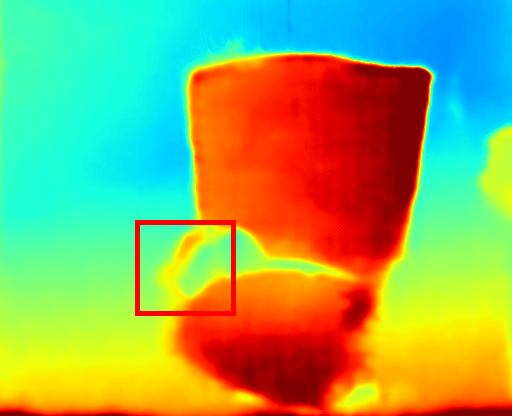} &
        \includegraphics[width=0.14\linewidth]{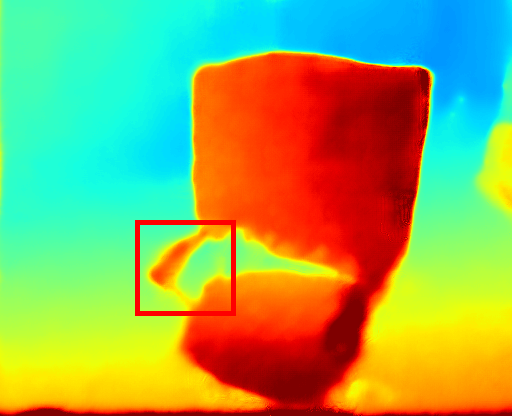} &
        \includegraphics[width=0.14\linewidth]{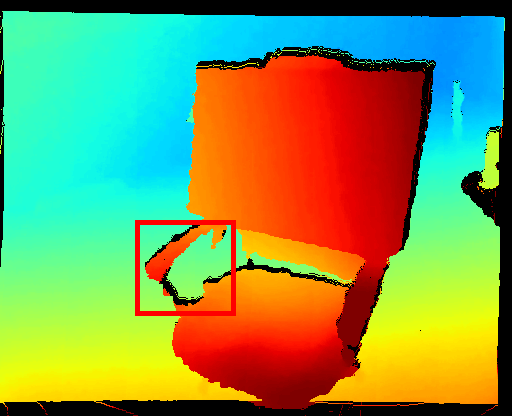}\\
        \includegraphics[width=0.14\linewidth]{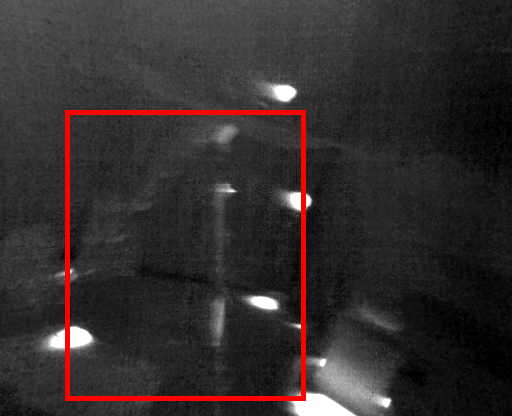} & 
        \includegraphics[width=0.14\linewidth]{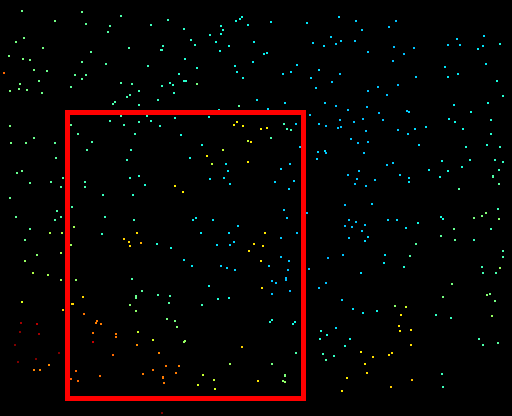} & 
        \includegraphics[width=0.14\linewidth]{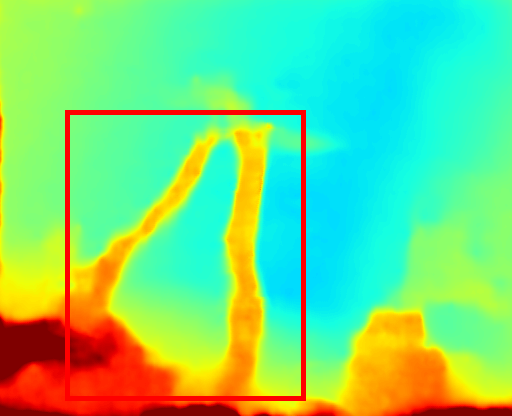} & 
        \includegraphics[width=0.14\linewidth]{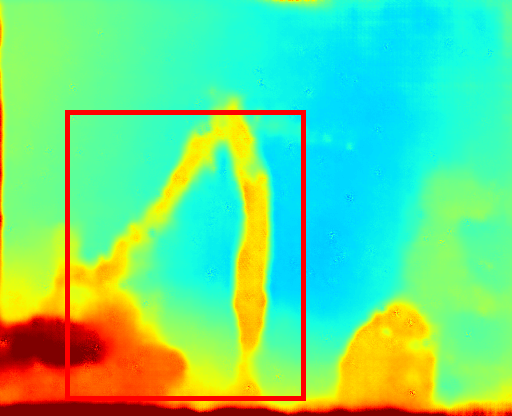} &
        \includegraphics[width=0.14\linewidth]{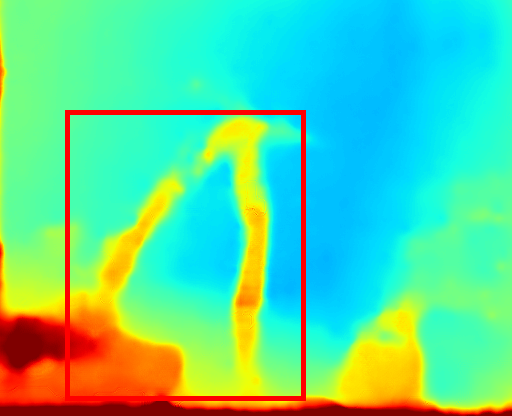} &
        \includegraphics[width=0.14\linewidth]{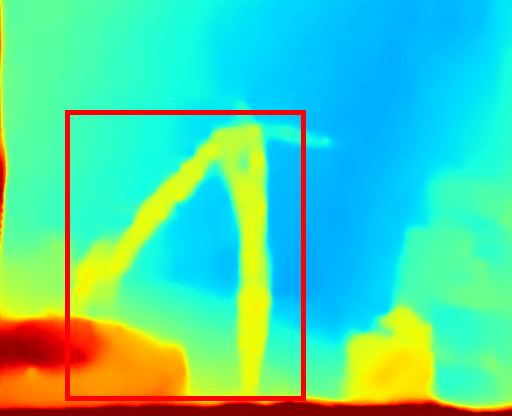} &
        \includegraphics[width=0.14\linewidth]{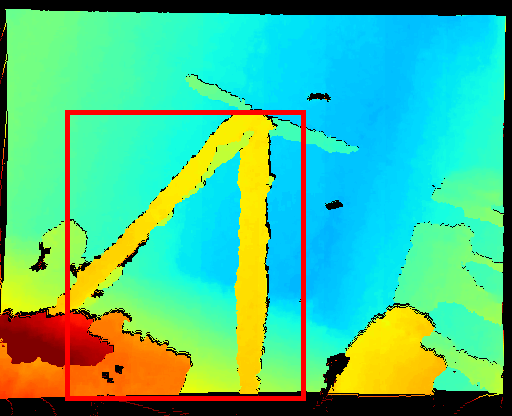}\\
        \footnotesize Thermal & \footnotesize Sparse depth & \footnotesize NLSPN~\cite{park2020non} & \footnotesize GuideNet~\cite{tang2020learning} & \footnotesize LRRU~\cite{wang2023lrru} & \footnotesize LRRU + Ours & \footnotesize GT
    \end{tabular}
    \caption{
    \textbf{Depth map comparisons on the ViViD dataset~\cite{lee2022vivid++} in dark scenarios.}
    Note that we dilated the sparse depth map.}
\label{fig:ViViD_visual_compare}
\end{figure*}

\begin{figure}[t]
\centering    
\renewcommand{\arraystretch}{0.2}
\begin{tabular}{c@{\hskip 0.0015\linewidth}c@{\hskip 0.0015\linewidth}c@{\hskip 0.0015\linewidth}c}
    \includegraphics[width=0.24\linewidth]{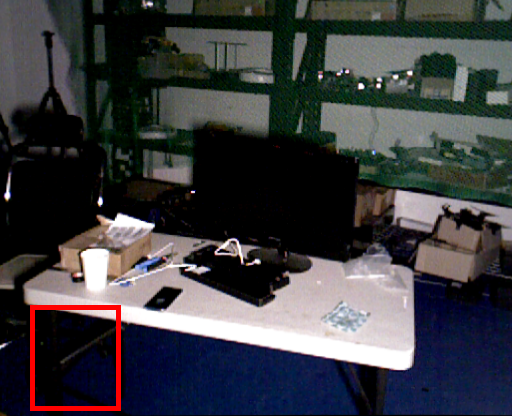} & 
    \includegraphics[width=0.24\linewidth]{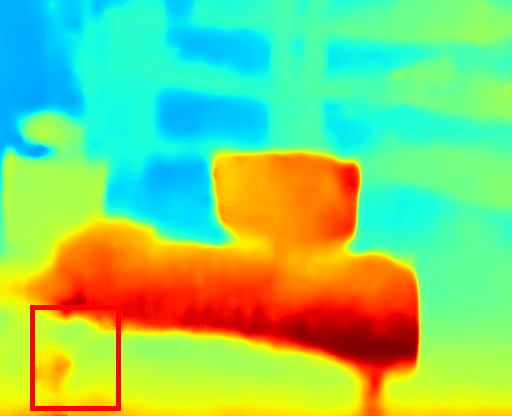} & 
    \includegraphics[width=0.24\linewidth]{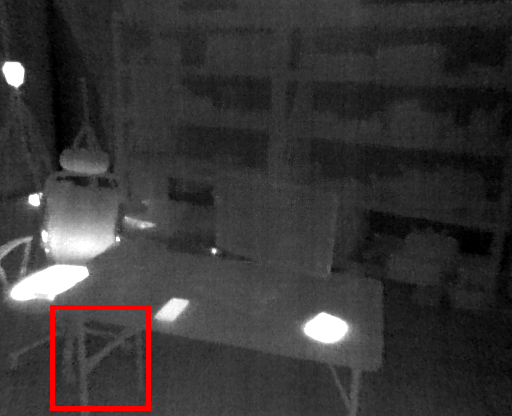} & 
    \includegraphics[width=0.24\linewidth]{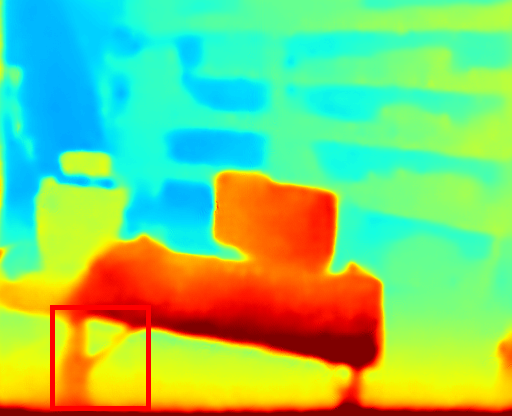}
    \\
    \includegraphics[width=0.24\linewidth]{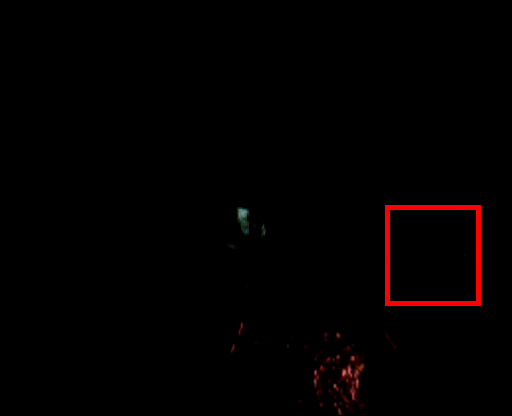} & 
    \includegraphics[width=0.24\linewidth]{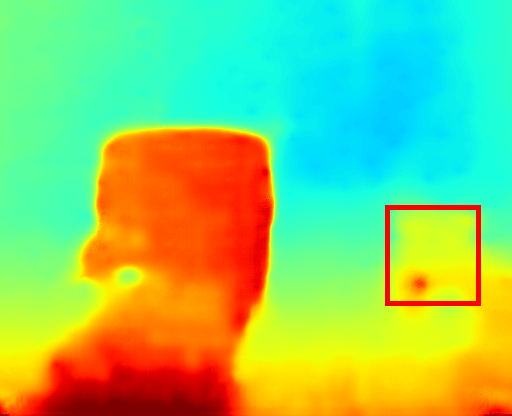} & 
    \includegraphics[width=0.24\linewidth]{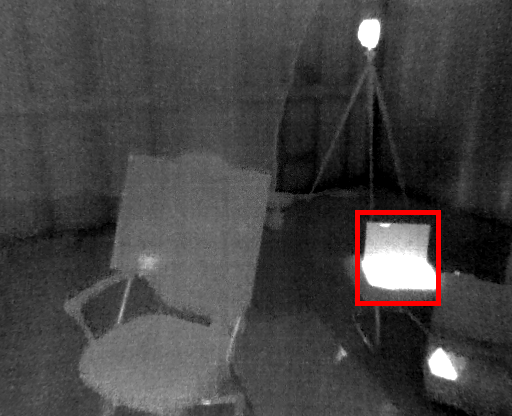} & 
    \includegraphics[width=0.24\linewidth]{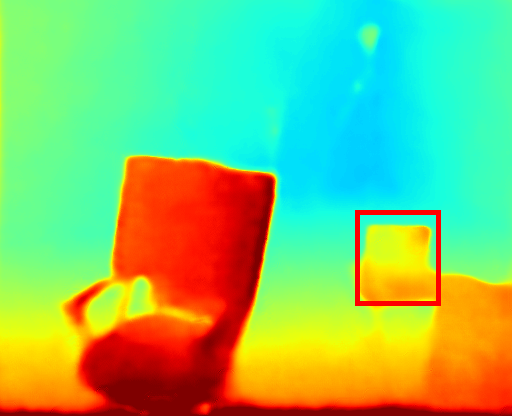} \\
    \footnotesize RGB & \footnotesize LRRU (RGB) & \footnotesize Thermal & \footnotesize LRRU (Thermal)
\end{tabular}
\centering
\caption{\textbf{Depth map comparisons between two modalities on the LRRU~\cite{wang2023lrru}.}
The first row presents the results of bright scenarios, while the second row corresponds to dark scenarios.}
\label{fig:vivid_modality_compare}
\end{figure}

\subsection{ViViD Dataset}
\label{subsec:ViViD Dataset}
\Tabref{tab:vivid_result}  presents performance comparisons of depth completion algorithms on the ViViD dataset across RGB and thermal domains, demonstrating the importance of thermal imaging for indoor scenes.
Due to the extremely low visibility in the ViViD dataset, optimizing depth prediction remains challenging in the RGB domain. This limitation results in significantly lower accuracy compared to the thermal domain across all depth completion networks.
The performance difference between indoor-bright and indoor-dark scenarios remains minimal in the thermal domain, with RMSE variations of less than 0.01 across all networks. In contrast, the RGB domain shows significant performance gaps in low-light conditions, particularly for networks such as NLSPN~\cite{park2020non}, GuideNet~\cite{tang2020learning}, and S2D~\cite{ma2018sparse}, where RMSE differences exceed 0.05 between the two scenarios.
This observation suggests that RGB-based depth completion is significantly impacted by low-light indoor environments, leading to unstable learning and degraded performance.
Moreover, image-guided methods without SPN (\eg, S2D and GuideNet) demonstrate a significant performance gap between the two modalities, highlighting the limitations of RGB as an effective guidance signal. This further emphasizes the critical role of thermal images in depth completion, as RGB images struggle to provide reliable geometric information in such low-light conditions.
\Figref{fig:vivid_modality_compare} highlights the advantages of thermal imaging in depth prediction.
Even in well-lit indoor environments, certain areas may remain shaded or visually indistinct in RGB images. However, thermal-based depth prediction effectively captures fine details of small objects, such as table legs, which might otherwise be overlooked.
Furthermore, thermal imaging enhances depth estimation for objects with strong heat signatures, enabling the prediction of depth for objects that RGB-based methods fail to recognize.

From a performance perspective of networks, CompletionFormer~\cite{youmin2023completionformer} achieves state-of-the-art performance in both of two modalities due to their geometry-awareness from the transformer backbone.
In the thermal domain, LRRU~\cite{wang2023lrru} achieves the second-best MAE, while in the RGB domain, it ranks second in RMSE. Its performance aligns well with the trend observed on the $\text{MS}^2$ dataset, demonstrating its consistency across different benchmarks.
We exclude BP-Net~\cite{tang2024bilateral} from this evaluation as it cannot perform the pre-processing step required to generate first stage depth map under highly sparse depth input conditions.

\Tabref{tab:ours} presents the effectiveness of our supervision approach using thermal images.
Our proposed method consistently improves the performance of NLSPN, GuideNet, and LRRU networks, as measured by RMSE and MAE.
Among these, the most significant improvement observed in LRRU, also demonstrated on the $\text{MS}^{2}$ dataset. These results suggest that supervision from the thermal depth foundation model effectively provides additional depth priors, while the short-to-long-range propagation strategy successfully captures dense depth information across various regions.
Although the performance gain for NLSPN appears marginal, our method applied to NLSPN outperforms the base supervision on LRRU in terms of RMSE, while achieving same accuracy in MAE. Importantly, these improvements are achieved without additional computational cost, as our method only requires supervision with pseudo-depth map from thermal image.

Furthermore, the effectiveness of our approach in capturing small objects in indoor scenarios with low visibility is validated, as demonstrated in \figref{fig:ViViD_visual_compare}. The highlighted areas show our method's strength in detail preservation.
Specifically, tiny objects that are missing even in the ground truth data are effectively reconstructed through direct supervision and sorted using similar pseudo-depth values via contrastive learning near depth boundaries.
These capabilities are critical for reducing misperception of small objects and avoiding false detection of non-object areas as short-distance regions, thereby enhancing the reliability of depth completion in low-light environments.
\subsection{Ablation Study}
\label{subsec:ablation_study}
We have conducted ablation studies using NLSPN~\cite{park2020non}. For the individual ablation experiments of depth-aware contrastive learning and pseudo-depth supervision, we employed a ResNet-18 backbone within NLSPN to facilitate faster experimental comparisons.
For the combined approach, we used the standard NLSPN model with a ResNet-34 backbone, which is its default configuration.

\begin{table}[t]
\caption{Performance comparisons of the index number $n$ and maximum margin $\psi_{max}$ in depth-aware contrastive learning.}
\centering
\footnotesize
\resizebox{0.82\linewidth}{!}{
\begin{tabular}{c|c|c|c|c}
\Xhline{3\arrayrulewidth}
\multicolumn{3}{c|}{Method} & \multicolumn{2}{c}{Metrics} \\ \hline
\multirow{2}{*}{Loss objectives} & \multirow{2}{*}{$n$} & \multirow{2}{*}{$\psi_{max}$} & RMSE  & MAE \\
 & & & (m)  & (m) \\ \hline\hline
$L_{base}$ & - & - & 2.504  & 1.441 \\ \hline
\multirow{7}{*}{$L_{base} + L_{contr}$} & 45 & - & 2.507 & 1.467 \\
& 60 & - & 2.434 & 1.374 \\
& 75 & - & 2.480 & 1.421 \\
& 90 & - & 2.454 & 1.428 \\ 
& 60 & $10$ & \underline{2.422} & \underline{1.363}  \\ 
& 60 & $20$ & \textbf{2.411} & \textbf{1.354}  \\ 
& 60 & $30$ & 2.426 & 1.370  \\ 
\Xhline{3\arrayrulewidth}
\multicolumn{5}{r}{\scriptsize{\textbf{Bold}: The best, \underline{Underline}: The second-best}} \\
\end{tabular}}
\label{tab:ablation_contr}
\end{table}

\subsubsection{Depth-aware Contrastive Learning}
To evaluate the effectiveness of our depth-aware contrastive learning approach, we conducted a detailed investigation of the sampling number for depth slicing indices and the margin sampling range, as shown in \tabref{tab:ablation_contr}. 
Initially, we analyze the impact of the sampling number $n$ on the performance of contrastive learning. Various values of $n$ are systematically tested to assess their influence on feature discrimination.
When $n$ is set to 45, the model shows no improvement in feature discrimination compared to the baseline model ($L_{base}$). This indicates that insufficient sampling limits the ability to effectively separate features in the representation space.
Through experimental analysis, we identify $n = 60$ as the optimal sampling number, yielding performance improvements of 2.8\% and 4.7\% in terms of RMSE and MAE, respectively.

Subsequently, we fix $n = 60$ and conduct further experiments to determine the best negative margin range. All tested ranges, including $\psi_{max} = 10, 20, 30$, demonstrate performance gains compared to the approach without a negative margin.
This result suggests that our margin sampling strategy, which selects only relatively negative samples based on pseudo-depth values, effectively enhances contrastive learning.
Among these, we observe that $\psi_{max} = 20$ provides the most suitable range, which allows effective contrastive learning.
This configuration achieves performance improvements of 3.7\% in RMSE and 6.0\% in MAE, effectively organizing the feature representation space while maintaining computational efficiency.
Therefore, we adopt $n = 60$ and $\psi_{max} = 20$ as the final settings for depth-aware contrastive learning in other experiments.

\subsubsection{Pseudo-depth Supervision}

\begin{table}[t]
\caption{Performance comparison of random sampling ratio $\alpha$ of $\mathcal{R}_{\text{sample}}$ in pseudo-depth supervision.}
\centering
\footnotesize
\resizebox{0.70\linewidth}{!}{
\begin{tabular}{c|c|c|c}
\Xhline{3\arrayrulewidth}
\multicolumn{2}{c|}{Method} & \multicolumn{2}{c}{Metrics} \\ \hline
\multirow{2}{*}{Loss objectives} & \multirow{2}{*}{$\alpha$} & RMSE  & MAE \\
 &  & (m)  & (m) \\ \hline\hline
$L_{base}$ & - & 2.504 & 1.441 \\ \hline
\multirow{6}{*}{$L_{base} + L_{SI}$} & - & 2.475 & 1.409  \\
& $0.10$ & 2.437 & 1.370  \\ 
& $0.15$ & 2.434 & 1.375  \\ 
& $0.20$ & \underline{2.418} & \textbf{1.359}  \\ 
& $0.25$ & \textbf{2.414} & \underline{1.361} \\ 
& $0.30$ & 2.427 & 1.375 \\ 
\Xhline{3\arrayrulewidth}
\multicolumn{4}{r}{\scriptsize{\textbf{Bold}: The best, \underline{Underline}: The second-best}} \\
\end{tabular}}
\label{tab:ablation_self-supervision}
\end{table}

We evaluated the impact of our random sampling strategy for pseudo-depth supervision, as shown in \cref{tab:ablation_self-supervision}.
While the full supervision approach yields only a 1.2\% improvement in RMSE, the random sampling method achieves over 2.7\% performance enhancement in RMSE across all tested sampling ratios ($\alpha$).
We observe consistent performance gains as $\alpha$ increases from 0.10 to 0.25, indicating that a moderately larger sampling ratio benefits the model by providing more diverse supervision.
These results highlight the effectiveness of introducing randomness into pseudo-depth supervision, enabling the model to focus on a broader set of features rather than overfitting to specific regions.
However, beyond $\alpha = 0.30$, performance begins to decline, suggesting that excessive sampling may introduce noise or weaken the model's ability to focus on meaningful depth information alongside ground truth data.
Among the tested ratios, the random sampling with $\alpha = 0.25$ achieves substantial performance improvements of 3.6\% and 5.6\% in RMSE and MAE, respectively.
Thus, we adopt a 25\% sampling ratio for the direct supervision approach using the scale-invariant loss function in subsequent experiments.
By ensuring globally unbiased sampling of pseudo-depth points, the random sampling strategy prevents the model from disproportionately focusing on densely supervised areas, fostering a more balanced understanding of the scene.

\subsubsection{Effectiveness of the Proposed Framework}

\begin{table}[t]
\caption{Performance comparison of various combinations of our proposed methods on the NLSPN~\cite{park2020non} network.
}
\footnotesize
\centering
\resizebox{0.90\linewidth}{!}{
\begin{tabular}{c|c|c|c|c}
\Xhline{3\arrayrulewidth}
Method & \multirow{2}{*}{$L_{contr}$} & \multirow{2}{*}{$L_{SI}$} & RMSE  & MAE \\
($L_{base}$) & & & (m) & (m) \\\hline\hline
\multirow{4}{*}{NLSPN (ResNet-34)} 
&  &  &  2.361 & 1.333 \\
& \checkmark & & \underline{2.351} & \underline{1.329} \\
&  &  \checkmark & 2.353 & 1.333 \\
& \checkmark & \checkmark & \textbf{2.347} & \textbf{1.328} \\
\Xhline{3\arrayrulewidth}
\multicolumn{5
}{r}{\scriptsize{\textbf{Bold}: The best, \underline{Underline}: The second-best}} \\
\end{tabular}}
\label{tab:ablation_methods}
\end{table}

\Tabref{tab:ablation_methods} demonstrates the effectiveness of our proposed methods. Both depth-aware contrastive learning ($L_{contr}$) and pseudo-depth supervision ($L_{SI}$) consistently improve performance on the $\text{MS}^2$ dataset.
Notably, integrating the contrastive learning approach into the NLSPN network with the ResNet-34 backbone results in a performance gain, reducing RMSE from 2.361 to 2.351 and MAE from 1.333 to 1.329.
Utilizing direct self-supervision with scale-invariant loss also reduces both of RMSE and MAE metrics.
Moreover, the proposed stage-wise learning strategy ($L_{pseudo}$) in \cref{eq:pseudo_loss} effectively stabilizes training and improves performance.
In particular, the stage-wise strategy decouples the scale-invariant loss and contrastive loss, enabling the model to first focus on global consistency and then refine local depth relationships. This sequential learning ensures a more stable training process and allows the model to effectively capture both large-scale and fine-grained depth details, all within a single training procedure.
As a result, combining both supervision methods with the pseudo-depth map further improves RMSE from 2.361 to 2.347.
Regarding the balance between two metrics, the final supervision strategy is established as a combination of $L_{contr}$ and $L_{SI}$ through stage-wise learning.
\section{Discussion}
While thermal depth completion offers several advantages, it still faces unresolved limitations.
Thermal images inherently suffer from low resolution, blurriness, and noise, particularly in complex scenes.
Additionally, depth prediction is highly influenced by the thermal emissivity of materials, leading to inconsistencies in reflective or transparent surfaces. The lack of fine texture details and low contrast further limit the effectiveness of conventional self-supervised constraints.

To address these challenges, future research should explore adaptive pseudo-depth supervision, where dynamically estimated confidence maps refine depth predictions based on the reliability of thermal features.
Such an approach would facilitate more robust fusion with auxiliary depth information, effectively mitigating noise and compensating for emissivity-related inconsistencies.
Moreover, environmental factors such as humidity and extreme temperatures introduce thermal noise, further degrading depth prediction.

Developing adaptive processing techniques and physics-based thermal modeling will be crucial to improve the robustness under varying conditions.
Another key challenge is the inherent sparsity of the LiDAR data, which becomes more pronounced in adverse weather conditions such as heavy rain, snow, or fog due to signal attenuation.
To mitigate this, future work should explore adaptive filtering techniques and temporal information integration to enhance depth completion under sparse conditions.
Additionally, alternative sensor fusion strategies should be investigated. Radar sensors, which remain robust in poor weather, provide structural information that complements the temperature-based perception of thermal imaging. 
A multi-modal fusion framework integrating radar and thermal imaging has the potential to enhance depth completion reliability in extreme environments.

Furthermore, existing depth completion architectures are primarily designed for RGB and LiDAR modalities and may not fully exploit the unique characteristics of thermal imaging.
Developing modality-specialized architectures that incorporate physics-informed learning and multi-scale feature extraction can improve depth prediction by addressing emissivity-related inconsistencies and refining structural details.

Future research should prioritize advancements in self-supervised learning for thermal depth completion, LiDAR sparsity compensation, and robust sensor fusion strategies. Additionally, improvements in adaptive thermal image processing and specialized architectures will be essential for achieving reliable depth estimation in real-world applications.
\section{Conclusion}
In this paper, we have conducted a comprehensive evaluation of thermal-LiDAR depth completion that better reflects real-world environments, including nighttime, rainy conditions, and low-light scenarios.
Through extensive benchmarking on the MS$^{2}$~\cite{shin2023deep} and ViViD~\cite{lee2022vivid++} datasets, we highlight the limitations of RGB-based depth completion under adverse conditions and demonstrate the robustness of thermal imaging in maintaining reliable depth estimation.
Furthermore, we introduce supervision approach that leverages a depth foundation model to maximize the benefits of thermal imaging under challenging conditions.
To further enhance depth completion accuracy, we propose the COntrastive and Pseudo-Supervised learning (COPS) framework, which refines depth boundaries through depth-aware contrastive learning and mitigates incomplete supervision by using a depth foundation model.
Our proposed margin sampling strategy selectively prioritizes relevant depth variations, while a scale-invariant loss with random sampling ensures a more balanced supervision signal. The proposed approach consistently improves performance across various depth completion networks without introducing additional computational overhead during inference. 
We believe our findings provide valuable insights into the challenges of thermal depth completion and encourage future research for real-world applications.

%
%
{\small
\bibliographystyle{IEEEtran}
\bibliography{references}
}

\end{document}